\documentclass[acmsmall, nonacm]{acmart}
\usepackage[utf8]{inputenc}

\usepackage{xcolor}

\usepackage{amsmath}
\usepackage{mathtools}
\usepackage{caption}

\usepackage[ruled,noend]{algorithm2e}
\usepackage{algpseudocode}

\usepackage{graphicx}
\usepackage{pgfplots}
\pgfplotsset{compat=1.15}
\usepackage{pgfplotstable}
\usepackage{filecontents}
\usepgfplotslibrary{statistics}
\usepgfplotslibrary{patchplots}
\usepackage{csvsimple}
\usepackage{multicol}
\usepackage{multirow}

\usepackage{subcaption}

\newcommand{\calA}[0]{\ensuremath{\mathcal{A}}}
\newcommand{\calC}[0]{\ensuremath{\mathcal{C}}}
\newcommand{\calE}[0]{\ensuremath{\mathcal{E}}}
\newcommand{\calL}[0]{\ensuremath{\mathcal{L}}}
\newcommand{\calM}[0]{\ensuremath{\mathcal{M}}}
\newcommand{\calN}[0]{\ensuremath{\mathcal{N}}}
\newcommand{\calS}[0]{\ensuremath{\mathcal{S}}}
\newcommand{\calT}[0]{\ensuremath{\mathcal{T}}}
\newcommand{\calX}[0]{\ensuremath{\mathcal{X}}}
\newcommand{\calV}[0]{\ensuremath{\mathcal{V}}}

\newcommand{\argmax}[0]{\ensuremath{\operatorname{argmax}}}
\newcommand{\argmin}[0]{\ensuremath{\operatorname{argmin}}}

\usepackage[size=tiny]{todonotes}
\newcommand{\sidenote}[1]{\todo[color=yellow!10, linecolor=yellow!50!black]{#1}}

\usepackage{environ}
\newenvironment{copypasted}{\color{blue!50!black}}{\color{black}}
\NewEnviron{hide}{}

\pgfplotsset{
MediumBarPlot/.style={
    font=\small,
    ybar,
    width=\linewidth,
    ymin=0,
    xtick=data,
    xticklabel style={text width=2.5cm, align=center},
    xtick pos=left,
},
BlueBars/.style={
    fill=MidnightBlue!50, bar width=4.5
},
RedBars/.style={
    fill=red!40, bar width=4.5
}
}
\pgfplotsset{
        select coords between index/.style 2 args={
        x filter/.code={
            \ifnum\coordindex<#1\fi
            \ifnum\coordindex>#2\fi
        }
    },
    greedy/.style={
        xshift=-\pgfkeysvalueof{/pgfplots/greedy scale}
    },
    genalg/.style={
        xshift=-\pgfkeysvalueof{/pgfplots/genalg scale}
    },
    simanneal/.style={
        xshift=-\pgfkeysvalueof{/pgfplots/simanneal scale}
    },
    integer/.style={
        xshift=-\pgfkeysvalueof{/pgfplots/integer scale}
    },
    rshift/.style={
        xshift=\pgfkeysvalueof{/pgfplots/rshift scale}
    },
    lshift/.style={
        xshift=-\pgfkeysvalueof{/pgfplots/lshift scale}
    },
    rshift scale/.initial=0.2em,
    lshift scale/.initial=0.2em,
    greedy scale/.initial=0.5em,
    genalg scale/.initial=1em,
    simanneal scale/.initial=1.5em,
    integer scale/.initial=0em,
}

\pgfplotsset{grid style={dashed,gray}}
\pgfplotsset{minor grid style={dashed,red}}
\pgfplotsset{major grid style={dotted,green!50!black}}

\usepackage[capitalize]{cleveref}

\title{Artificial Intelligence for Smart Transportation}

\author{Michael Wilbur}
\affiliation{\institution{Vanderbilt University}
\country{USA}
}

\author{Amutheezan Sivagnanam}
\affiliation{\institution{Pennsylvania State University}
\country{USA}
}

\author{Afiya Ayman}
\affiliation{\institution{Pennsylvania State University}
\country{USA}
}

\author{Samitha Samaranayeke}
\affiliation{\institution{Cornell University}
\country{USA}
}

\author{Abhishek Dubey}
\affiliation{\institution{Vanderbilt University}
\country{USA}
}
\author{Aron Laszka}
\affiliation{\institution{Pennsylvania State University}
\country{USA}
}

\begin{document}

\setcounter{tocdepth}{2}
% \tableofcontents
% \clearpage
\setcounter{page}{1}

\maketitle

\section{Introduction}

There are more than 7,000 public transit agencies in  the U.S. (and many more private agencies), and together, they are responsible for serving 60 billion passenger miles each year.
Additionally, new on-demand modalities including ride-share, bike-share, and e-scooters have been introduced in recent years and transformed the transportation landscape in urban environments.
A well-functioning transit system fosters the growth and expansion of businesses, distributes social and economic benefits, and links the capabilities of community members, thereby enhancing what they can accomplish as a society~\cite{beyazit2011evaluating,harvey2010social,federal2003status}. 
However, the explosion in transportation options and the complicated relationship between public and private offerings present myriad new challenges in the design and operation of these systems.
There are also complex, and often competing, operational objectives that complicate the implementation of efficient services.
Since affordable public transit services are the backbones of many communities,
solving these problems and understanding state-of-the-art methods for AI-driven smart transportation has the potential to strengthen urban communities, address the climate challenge, and foster equitable growth.

Fundamentally, the design of a well-functioning transit system requires solving complex combinatorial optimization problems related to planning and real-time operations.
These problems span many well studied fields, from classical line planning to offline and online vehicle routing problems (VRPs).
While there are many ways to assess the performance of smart transportation systems, we largely focus on evaluating these systems in the context of optimizing \textit{utilization} (i.e. ridership) and \textit{efficiency} (i.e. reducing operational costs).
Increasing utilization requires learning mobility patterns over wide geographical areas 
and adapting systems to better meet the demand for mobility.
It also requires flexible mobility options such as on-demand and multi-modal transportation to adapt to demand in real-time and thus better serve potential passengers.
Additionally, more efficient systems alleviate the impact on the environment by reducing emissions and can free resources by reducing costs.
Lastly, \textit{coverage} is an important consideration that often competes with utilization and efficiency.
Therefore it is important to discuss optimization in the context of \textit{ridership} versus \textit{coverage}, the latter of which is an important consideration in the design of equitable and fair transportation.
While the problem of ridership versus coverage is open-ended, there are ways in which these dimensions can be modelled such that transit agencies can better optimize over the objectives that matter most to them.

Efficient transportation systems require making decisions in real-time over large geographical areas and computationally intractable state-action spaces.
Due to the intractability of these problems, traditional analytical methods fall short.
Therefore, transit agencies have turned to computational approaches that enable large-scale data-driven optimization.
AI-driven transportation can address this problem by learning complex abstractions of vast volumes of data to make decisions in a computationally efficient manner at scale and in real-time.
However, it requires solving complex challenges in model representation, decision-making under uncertainty, as well as data ingestion and processing.

%Designing efficient transportation systems involves solving intractably large problems in which traditional analytical methods fall short.
%Therefore, transit agencies have turned to computational approaches that enable large-scale data-driven optimization.

%In this way, AI algorithms can leverage complex abstractions of vast volumes of data to make decisions or model the environment in a computationally efficient manner that allows transit operators to solve problems at city-scale.
%However, there are numerous challenges associated with designing data-driven, or AI-driven transportation systems.
%First, transportation data is high-volume and high-velocity so there must be carefully designed data processing and ingestion methods for handling various structured and unstructured data.
%Second, smart transportation systems require making decisions in real-time over large geographies and computationally intractable state-action spaces.
%This requires computationally efficient environmental models and decision-making algorithms.

%Efficient transportation systems require making decisions in real-time over large geographies and computationally intractable state-action spaces.

Since smart transportation is a broad field,  we primarily focus on two high-impact modalities: fixed-line and on-demand transportation systems.
Fixed-line transportation utilizes % consists of a fixed schedule of 
vehicles (bus or rail) that traverse pre-defined routes over the course of a day following fixed schedules.
On-demand transportation is a flexible service in which users request rides either in real-time or ahead-of-time, and transportation is provided from origin to destination through ride-share, e-scooters, or bike-share.
Additionally, many transit agencies have recently investigated the potential of multi-modal transportation, which provides a more flexible and adaptive transportation model that combines fixed-line and on-demand transportation.

%This chapter outlines key challenges in these domains as well as the potential of multi-modal transportation for next generation transit which provides a more flexible and adaptive transportation model that combines fixed-line and on-demand transportation.

This chapter discusses the primary requirements, objectives, and challenges related to the design of AI-driven smart transportation systems.
The major content involves the following:
\begin{enumerate}
    \item Data sources and data management for AI-driven transportation.
    \item An overview of how AI can aid decision-making with a focus on transportation.
    \item Computational problems in the transportation domain and AI approaches.
\end{enumerate}

%A key component of all problems in this space is the balance between \textit{ridership} and \textit{coverage}. 
%Affordable public transit services are the backbones of many communities, providing diverse groups of people with access to employment, education, and other services.
%Affordable transit services are especially important in low-income neighborhoods where residents might not be able to afford personal vehicles.
%However, transit services that provide wide and equitable coverage tend to suffer from low utilization---compared to concentrating service into a few high-density areas---which leads to higher fuel usage per passenger per mile.
%This in turn results in higher operating costs, which  threatens affordability---a problem that has recently been exacerbated by the COVID-19 pandemic.

\section{Context}

\begin{figure}
\centering
  \includegraphics[width=.75\linewidth]{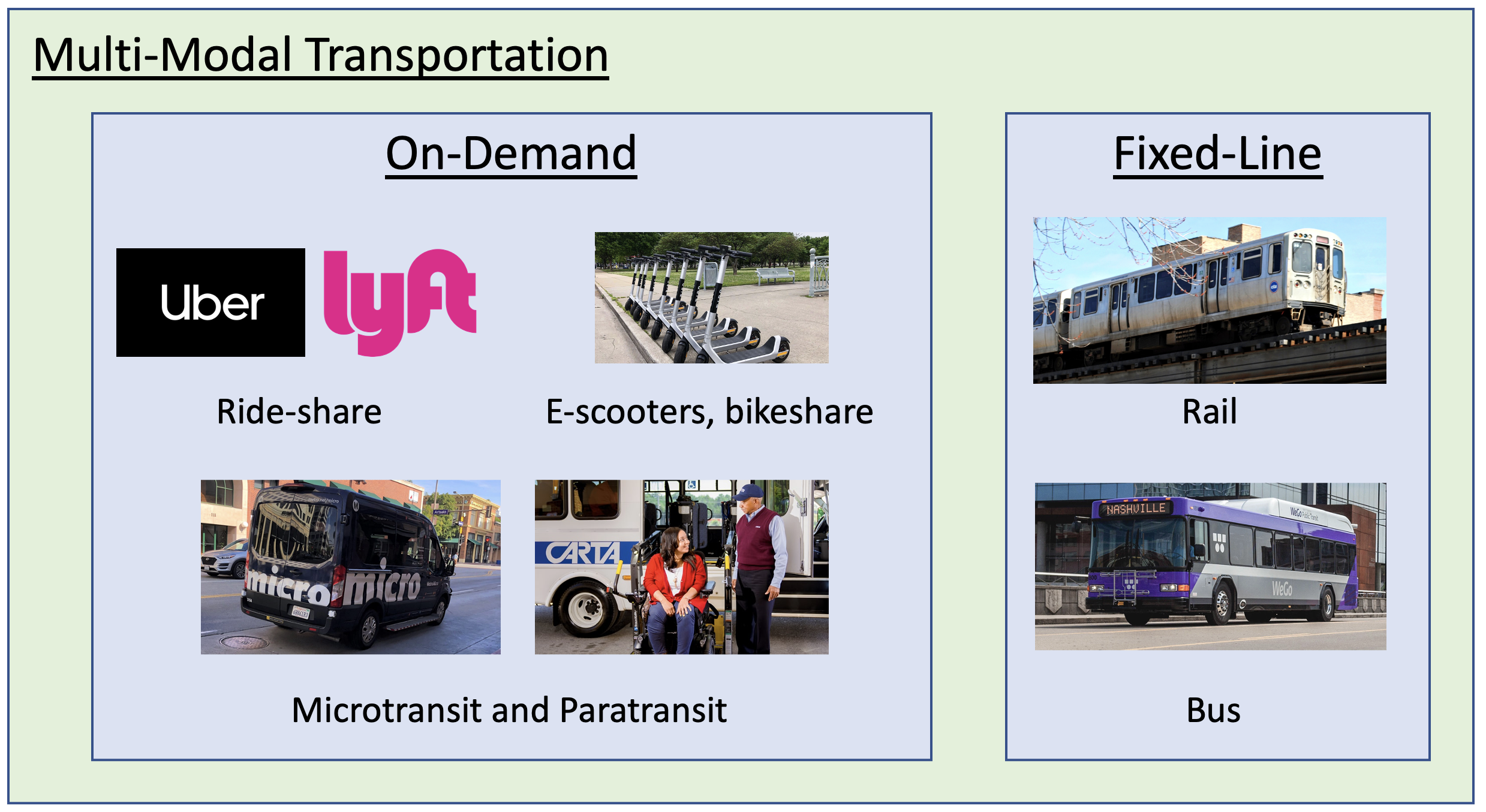}
  \caption{
  Transportation modalities: there is a variety of mobility options in today's urban environments, provided by both public transit agencies and private mobility providers, which can be classified as \textit{on-demand} or \textit{fixed-line}.
  %Transportation Modalities - Urban mobility consists of a heterogeneous mix of transportation options provided by both public transit agencies and private mobility providers which can be classified as \textit{on-demand} or \textit{fixed-line}. 
  On-demand transportation provides direct service and is often accessed through a smart-phone application. 
  Fixed-line transportation utilizes mass transit vehicles that service fixed routes and schedules.
  Multi-modal transportation is a broad categorization that includes any service that combines multiple modes of transit to serve passengers.}
  \label{fig:modalities}
\end{figure}

Large-scale adoption of smart phones and sensing technologies have revolutionized urban mobility in recent years.
Led by companies such as Uber, Lyft and Via, new on-demand transportation options such as ride-share, e-scooters, and bike-share have been introduced to complement existing transportation services provided by public transit agencies.
As the mobility landscape evolves, public transit agencies are tasked with managing ever more complex systems.
To understand the relationship between transportation modalities, we can classify these systems as \textit{on-demand}, \textit{fixed-line}, and \textit{multi-modal} as shown in Figure~\ref{fig:modalities}.

On-demand transportation provides direct service from origin to destination.
These systems are often accessed through a smart-phone application or by calling an agency to schedule trips.
Ride-share companies such as Uber and Lyft provide their own smart-phone applications for users to request rides in real-time.
Microtransit and paratransit provide similar services but are distinguished by the use of high-capacity vehicles.
In these systems, passengers typically share rides with several other users, and trip requests can be placed ahead-of-time (for example, a trip can be requested for later in the week at a specific time) or in real-time in the same way as ride-share.
These services can be provided by private mobility providers or be managed by a public transit agency.
From a city's perspective, microtransit services are available to all residents and can be thought of as a low-cost extension of their public transit system.
They can be used for direct point-to-point travel as well as in hybrid transit systems, where the vehicles shuttle passengers to and from fixed-line transit~\cite{salazar2018interaction}.
Similarly, paratransit provides curb-to-curb service for passengers who are unable to use fixed-route transit (e.g. passengers with disabilities) and is often considered a social good or required by law.
%As we will discuss, microtransit and paratransit are two ways in which public transit agencies are looking to 

Fixed-line transportation is the backbone of urban mobility across many urban centers.
These services are provided by public transit agencies and consist of large, mass-transit vehicles such as buses or rail-cars that service fixed routes and schedules.
A fixed route is a pre-defined sequence of locations that a vehicle will traverse with a scheduled arrival time at each location.
Transit agencies aim to provide service at each stop at relatively frequent intervals throughout the day.
Fixed-line services have high capacity and can transport passengers in a way that is more environmentally friendly compared to on-demand services or personal vehicles.
However, in many cities fixed-line transportation struggles to attract a critical mass of passengers while providing adequate coverage to all areas of the city \cite{wilbur2020impact}.
Therefore, optimizing these systems is important for increasing ridership and reducing costs in a way that properly serves the needs of passengers spread over wide geographical areas.

As a response to these challenges, public transit agencies have begun revamping their operations in a way that integrates on-demand services with fixed-line transit to provide dynamic coverage in areas where demand is spread out spatio-temporally, making it inefficient to serve these areas with high-capacity, fixed-line services.
We can refer to this as \textit{multi-modal} transportation.
There are many open challenges to implementing multi-modal transportation systems that are outside the scope of this work.
Instead, we focus on optimization methods for on-demand and fixed-line services.
\textit{Optimization in this context aims to improve ridership and efficiency while providing adequate service to residents across socio-economic groups.} 
By optimizing these services, we can better understand how these systems can be integrated more effectively.
The most promising way to optimize transportation services is through data collection and analysis.
Data enables transit agencies to improve their understanding of their systems to better inform decision-making as well as leverage recent advances in AI to optimize their services, which is the focus of this work.

\section{Background}

\begin{figure}
\centering
  \includegraphics[width=\linewidth]{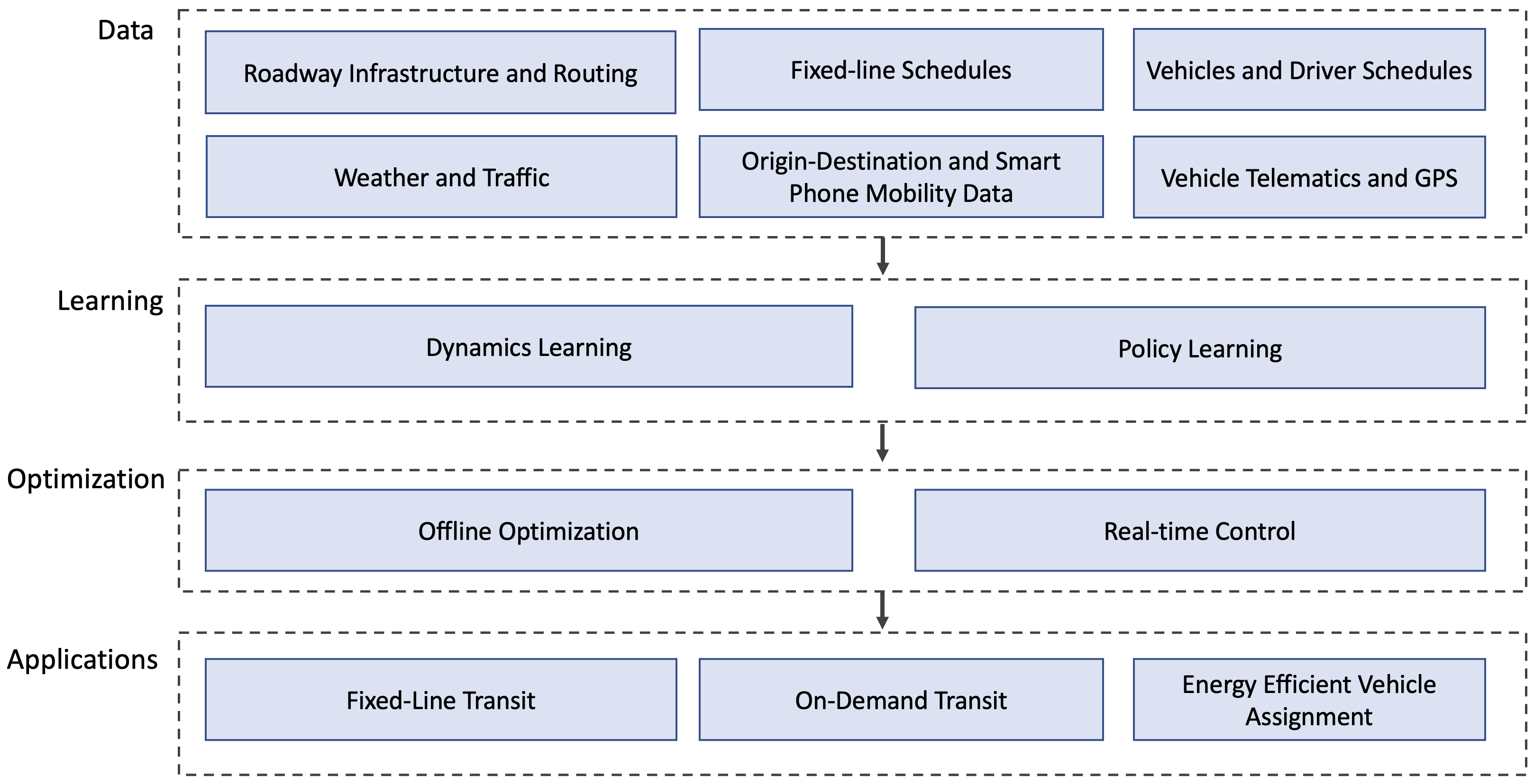}
  \caption{AI-driven smart transportation requires integrating data, learning, and optimization to provide smart transportation services and applications for transit agencies and users. These applications aim to optimize \textit{utilization} (i.e. increase ridership) and \textit{efficiency} (i.e. reduce operational costs and emissions).}
  \label{fig:overview}
\end{figure}

Effective solutions are of critical importance in that poorly designed mobility applications can make for unhappy users, increase costs, and in the case of safety critical applications, can affect the livelihood of residents that depend on the system.
The goal of these systems is to learn models from data that can aid optimization and support more efficient mobility options.
AI-driven smart transportation systems therefore require integrating \textit{data}, \textit{learning}, and \textit{optimization} as shown in Figure~\ref{fig:overview}.
Each one of these components presents unique challenges that must be addressed when designing smart transportation systems.
In this section, we summarize each of these components and how they can be combined to create smart transportation applications.
%First we start with an overview of transportation modalities.

\subsection{Data Sources and Sensing}

The ``data'' component refers to the need for high-quality data sources that can be used to generate datasets for training predictive models and to generate data online for real-time applications.
In the transportation domain, this data is mostly streamed from various sensors and is spatio-temporal in nature.
Therefore, sensor data typically includes a timestamp indicating when the reading was taken, the value of the reading, as well as a spatial component representing where this sensor is located.
Sensors can be statically located along roadways or can be dynamic in that they travel over time, e.g. GPS receiver within a vehicle or bus.
The spatio-temporal nature of these data sources presents challenges in efficient storage, synthesis, and data retrieval \cite{aydin2016modeling, wang2019integrated, wilbur2021efficient}. 
Synthesis requires processing data in a variety of domain-specific formats and at irregular intervals.
Raw sensor streams must often be enriched by joining them with infrastructure data, such as the identifier of the road along which a vehicle is travelling.

Data sources can be classified as static or real-time.
Static data are datasets related to the road networks, GIS layout, and schedules for the operating regions, while real-time data consists of data streaming from sensors and collected during real-time operations.
In this sense, the static data is not fully ``static.''
For instance, both roadway information and fixed-line schedules can change over time.
However, the static data in this context is not required to be updated daily in the way that real-time data is.
The most common forms of static data used for building smart transportation systems include the following.

%\subsubsection{Roadway Infrastructure and Routing Graphs}

\begin{itemize}
    \item \textit{Roadway Infrastructure and Routing Graphs:} Transportation models require a way to represent the roadway network.
    A commonly used roadway network data source is OpenStreetMap (OSM).
    OSM is community-driven and open-source with updates provided by the OSM community with a wide geographic footprint \cite{haklay2008openstreetmap}.
    Data can be directly downloaded from various mirrors \cite{geofabrik} and is used as the mapping data for a variety of commercial-grade applications.
    OSM in its raw form consists of \textit{nodes}, which are points with a geographic position (latitude, longitude); \textit{ways}, which are ordered lists of nodes representing linear features such as streets; and \textit{relations}, which define relationships between nodes and ways.
    Together, these features represent a graph that can be used for mapping or routing.
    There are numerous open-source projects that provide routing services such as shortest path calculation and travel time estimation using OSM data, including the Open Source Routing Machine (OSRM), OpenTripPlanner, and Valhalla \cite{osrm, valhalla}.
    \item \textit{Transit Schedules:} The most widely used schema for representing fixed-line routes and schedules is the General Transit Feed Specification (GTFS) \cite{staticgtfs}.
    GTFS describes a standard format for public transportation schedules and associated geographic information.
    This includes standard representations for routes, trips, stops, and scheduled arrival times at each stop for buses and subway networks.
    It also includes geo-spatial shape objects representing routes so that fixed-line transit can be mapped to the operating region.
    Public transit agencies maintain and publish their own schedules in GTFS format for public consumption.
    In this way, developers and researchers can write code that consumes GTFS data in an interoperable way that can work across transit agencies and regions, provided that the transit agencies maintain GTFS-compliant data sources.
    \item \textit{On-Demand Driver Runs:} Some forms of on-demand transportation, particularly those managed directly by public transit agencies, may include scheduling information for vehicles and drivers.
    For instance, in the case of micro-transit and para-transit, a transit agency maintains a list of available drivers for a day and the start and end times of their runs.
    It may also maintain the list of available vehicles including the capacity of each vehicle and which driver-run corresponds to each available vehicle.
    This contrasts with many commercial ride-share services, such as Uber or Lyft, where there are no fixed schedules for drivers. % or fixed vehicles each driver will use.
\end{itemize}

%\subsection{Environment Sensing - Real-time Data Sources}

Advances in sensing technologies have led to an explosion of real-time streaming data in urban environments.
These sensors allow us to monitor transportation systems in real-time and act as a bridge between the physical and cyber domains.
They allow us to quantify and monitor the system through data.
Real-time data then serves two purposes.
First, we can collect sensor data over a time range to generate offline datasets.
These offline datasets are joined with the static data and are used to train AI models.
Second, the learned models then ingest the sensor data at inference time to make predictions and aid  decision-making processes.

The data itself comes in a variety of formats and is typically defined as high-velocity and high-volume.
Therefore, the problem of ingesting, processing, storing, and interpreting real-time sensor data in the transportation domain inherits the traditional problems associated with Internet of Things (IoT) analytics \cite{mohammadi2018deep}.
Commonly used real-time data sources in the transportation domain are as follows.
%In this section we outline commonly used real-time data sources in the transportation domain.

\begin{itemize}
    \item \textit{Vehicle Telematics and GPS:} Vehicle telematics typically consist of physical sensing kits installed in vehicles.
    These devices can record vehicle dynamics such as speed, acceleration, and tractive or brake torque of a vehicle or flow of vehicles \cite{chen2020review}.
    They can also record power consumption and generation which can vary between types of vehicles (e.g. electric, diesel, and hybrid vehicles).
    AI-driven smart transportation relies heavily on vehicle positioning and trajectory tracking using GPS.
    GPS readings are streamed in real-time either through GPS-specific devices installed in the vehicle or through mobile applications on smart phones and tablets.
    In addition to monitoring vehicle locations, GPS trajectories can also be tracked for transit users through user-defined smart-phone applications.
    \item \textit{Origin-Destination and Smart-Phone Data:} Many optimization applications for smart transportation require origin-destination (OD) datasets which encompass trips for users throughout the day.
    At a minimum, OD data requires an origin location (where the user originates) and a destination location (where the user ends) for each trip.
    Optionally these locations can be timestamped in a way that represents when the user started and or ended the trip.
    Historical data may also include trajectories which represent geo-points along the trip for the transit user.
    One publically available source of OD-data is the Longitudinal Employer-Household Dynamics (LEHD) Origin-Destination Employment Statistics (LODES), which is provided by the United States Census Bureau \cite{lodesdata}.
    LODES data includes information related to census groups and tracts where residents live and work, which can be used to investigate demographics and trips between spatial regions within a city or state.
    There are also commercially available datasets derived from anonymized smart-phone GPS data, which can aid OD generation by provided traffic related to points-of-interest (POIs) \cite{safegraph}.
    \item \textit{GTFS Realtime:} GTFS is an open data format used by many transit agencies to represent public transportation schedules and associated geographic information.
    GTFS Realtime is an extension to the static GTFS schedules that provides up-to-date information about current arrival times, vehicle locations, occupancy, and service alerts to help users better plan their trips \cite{realtimegtfs}.
    The structured nature of GTFS Realtime makes it easy to monitor fixed-line public transit performance in the context of the transit schedules and routes.
    GTFS Realtime has been incorporated into a variety of commercial-grade applications including Google Maps.
    GTFS Realtime  is typically provided by transit agencies themselves.
    Compared to raw GPS data, GTFS Realtime is a structured format that is interoperable between agencies and telematics kit providers.
    \item \textit{Weather:} Weather conditions can have a significant impact on transportation systems including on energy usage and travel delays \cite{ecml2021}. 
    Raw weather data can be accessed for weather stations throughout the United States from the National Oceanic and Atmospheric Administration (NOAA).
    Common weather features include temperature, precipitation, atmospheric pressure, wind speed, and wind direction.
    Various APIs, such as Meteostat \cite{meteostat}, provide easy access to historical as well as real-time NOAA weather data for simple integration into AI pipelines.
    \item \textit{Traffic:} Traffic congestion and road segment speeds are important features for estimating travel time, routing, and energy prediction in smart transportation applications.
    Typically these datasets include real-time travel speeds or travel time along roadway segments that must be tied in some way to the roadway infrastructure.
    HERE Technologies provides an API for speed recordings for segments of major roads in the form of timestamped speed recordings \cite{heredata}.
    Every road segment is identified by a unique Traffic Message Channel identifier (TMC ID). Each TMC ID is also associated with a list of latitude and longitude coordinates, which describe the geometry of the road segment.
    Additionally, INRIX \cite{inrixdata} provides TMC-level traffic information that also is mapped to OSM segments and is therefore easily integrated with OSM-based applications.
\end{itemize}

\subsection{Optimization for Transportation}\label{subsection:optimization for transportation}

Smart transportation systems require solving computational problems that optimize various objectives, such as ridership, costs, or emissions over time.
Optimization problems in this setting can be broadly classified in two categories: offline and real-time.
In this section, we will discuss these two settings and the role that learning plays in solving these computationally challenging problems.

\subsubsection{Offline optimization problems}
Offline optimization problems are often solved ahead of time in a batch setting, and they are common in logistics and supply-chain optimization.
For example, a package delivery company may need to assign routes to drivers to deliver items the next day.
This class of problems is fundamentally combinatorial optimization.
Broadly, combinatorial optimization involves finding an optimal solution from a set of feasible solutions, where the set of feasible solutions is discrete or can be reduced to a discrete set; many of these problems are NP-hard.
In transportation settings, there are often strict constraints that must be satisfied.
For example, in the case of package delivery, there may be certain times of day when the package must arrive at a specific location or a fixed number of vehicles that are available.
Therefore, a feasible solution is a solution that satisfies all of the constraints, and
an optimal solution is a feasible solution that maximizes the objective.
Combinatorial optimization can be modelled as an integer linear program or mixed-integer linear program (MILP) and solved using state-of-the-art heuristic search approaches, such as branch-and-bound or branch-and-cut.
While these methods are exact in that they are guaranteed to converge to optimal solutions, they unfortunately reduce to potentially enumerating all feasible solutions, which poses scalability issues.

\subsubsection{Real-time control (online optimization problems)}\label{subsubsection: real-time control}
Conversely, real-time control (or online planning) refers to optimization problems where decisions must be made sequentially over time.
Sequential decision-making is commonly formalized as a Markov Decision Process (MDP) \cite{puterman2014markov}.
%These problems are stochastic and can be modelled as a sequential decision-making problem. 
An MDP can be specified by a tuple $\{\mathcal{S}, \mathcal{A}, \mathcal{T}, \mathcal{R}, p(s_o)\}$:
\begin{itemize}
    \item Set of states $\mathcal{S}$ and distribution over the starting state $p(s_0)$.
    \item Set of actions $\mathcal{A}$.
    \item Transition function $\mathcal{T}: \mathcal{S} \times \mathcal{A} \rightarrow \mathcal{S}$ takes as input the current state $s_t \in \mathcal{S}$ and an action $a_t \in \mathcal{A}$ at timestep $t$ and returns the next state $s_{t+1} \in \mathcal{S}$.
    %\item Transition function $\mathcal{T}(s_{t+1}|s_t, a_t)$ where $s_t$ is the state $s \in \mathcal{S}$ at timestep $t$ and action $a_t \in \mathcal{A}$ at timestep $t$.
    \item Reward function $\mathcal{R}: \mathcal{S} \times \mathcal{A} \times \mathcal{S \rightarrow \mathbb{R}}$.
    %\item Reward function $\mathcal{R}(s_t, a_t, s_{t+1})$.
    \item The policy $\pi: \mathcal{S} \rightarrow \mathcal{A}$ controls the agent's behaviour by selecting an action given the current state. 
\end{itemize}
The \textit{environment} consists of the transition function and reward function.
At each timestep $t$ the environment takes $s_t$, $a_t$ and returns the next state from $\mathcal{T}(s_{t+1}|s_t, a_t)$ and a scalar reward from $\mathcal{R}(s_t, a_t)$.
The objective of sequential decision-making is to select actions given a policy $\pi(a_t|s_t)$ to optimize \textit{cumulative} reward over time.

\subsection{Learning}

The fundamental advantage of learning-based AI methods is their ability to handle high-dimensional state-space and to \textit{learn} decision procedures/control algorithms from data rather than from models.
This is advantageous in the transportation domain, where real-world state-action spaces are high-dimensional and manifest as intractable state-spaces for mathematical modeling and analysis.
Therefore, we can employ state-of-the-art AI methods to learn abstract representations of these problems to tackle real-world transportation problems.
In this section, we discuss two broad categories of learned models: \textit{descriptive} models and \textit{generative} models.

%Before we discuss how AI and machine learning can drive optimization in transportation we first discuss two broad categories of learned models - \textit{descriptive} models and \textit{generative} models.
%An understanding of these two classes of models is important for studying AI-driven optimization in Section~\ref{subsection: ai driven optimization}.

%The fundamental advantage of AI methods is their ability to handle high-dimensional state-space and learn decision procedures/control algorithms from data rather than models. This is because high-dimensional real-world state spaces are complex and manifest as intractable state-spaces for mathematical modeling and analysis. Data-driven decision procedures can learn abstract representations of these state spaces and then applied in real-world CPS such as autonomous vehicles, proactive emergency response systems, transit management systems and electric power grids. 

%Due to the complexity of smart transportation systems, traditional analytical methods are infeasible.AI 

%The goal of the "learning" component is to learn various models from data that can aid downstream optimization components.

\subsubsection{Descriptive Models} Descriptive models utilize supervised learning to learn abstract representations of the model environment and are often referred to in the AI literature as discriminative models.
The models are trained with labeled data and aim to capture the conditional probability $p(Y|X)$ for input features $X$ and output labels $Y$.
That is, predictive models aim to learn a function $f: X \rightarrow Y$ that maps the input features to output labels.
%That is if the input features are $X$ and the output is $Y$, predictive models aim to capture the conditional probability $p(Y|X)$.
Supervised learning can be separated into two types of problems (or \textit{tasks}): classification and regression.
Classification tasks aim learn a function $f$ that maps the input features to labels that are discrete categories.
For example, given the features of a transit route and the current state-of-charge for an electric bus, we may want to predict whether the vehicle can service the route without having to re-charge.
This would be a binary classification problem.
On the other hand, regression tasks map the input features to a continuous label space.
For example, given the features of a transit route and the current state-of-charge for an electric bus, we may simply want to predict the energy that will be used for this vehicle to service the route.
There are many state-of-the-art methods for learning predictive models, including deep neural networks (DNN).

\subsubsection{Generative Models} On the other hand, {generative} models are a class of statistical models that learn the distribution of the environment $P(X)$, or in tasks where labelled data is available, learn the joint probability of $P(X, Y)$.
Generative models provide a condensed representation of the environment for building simulators, learning policies through reinforcement learning, or being queried at inference time to simulate future scenarios.
Generative models in the transportation domain are particularly important for modelling demand.
For example, we might create a generative demand model for representing the spatio-temporal distribution of trip requests for on-demand transportation, which can aid decision-making when assigning vehicles to service new requests in real-time.

%\subsection{Optimization}\label{subsection:optimization}

\subsection{AI-Driven Optimization}\label{subsection: ai driven optimization}

%MDPs can be deterministic when the transition function always returns the same next state for a given state-action pair or stochastic. 
%In a stochastic MDP the transition function instead returns a probability distribution over the possible next states.

%We can either approximate the entire next state distribution (descriptive models), or approximate a model from which we can only draw samples (generative model). Descriptive models are mostly feasible in small state spaces. Examples include tabular models, Gaussian models (Deisenroth and Rasmussen, 2011) and Gaussian mixture models (KhansariZadeh and Billard, 2011), where the mixture contribution typically involved expectationmaximization (EM) style inference (Ghahramani and Roweis, 1999). However, these methods do not scale well to high-dimensional state spaces.

\begin{figure}
\centering
  \includegraphics[width=.8\linewidth]{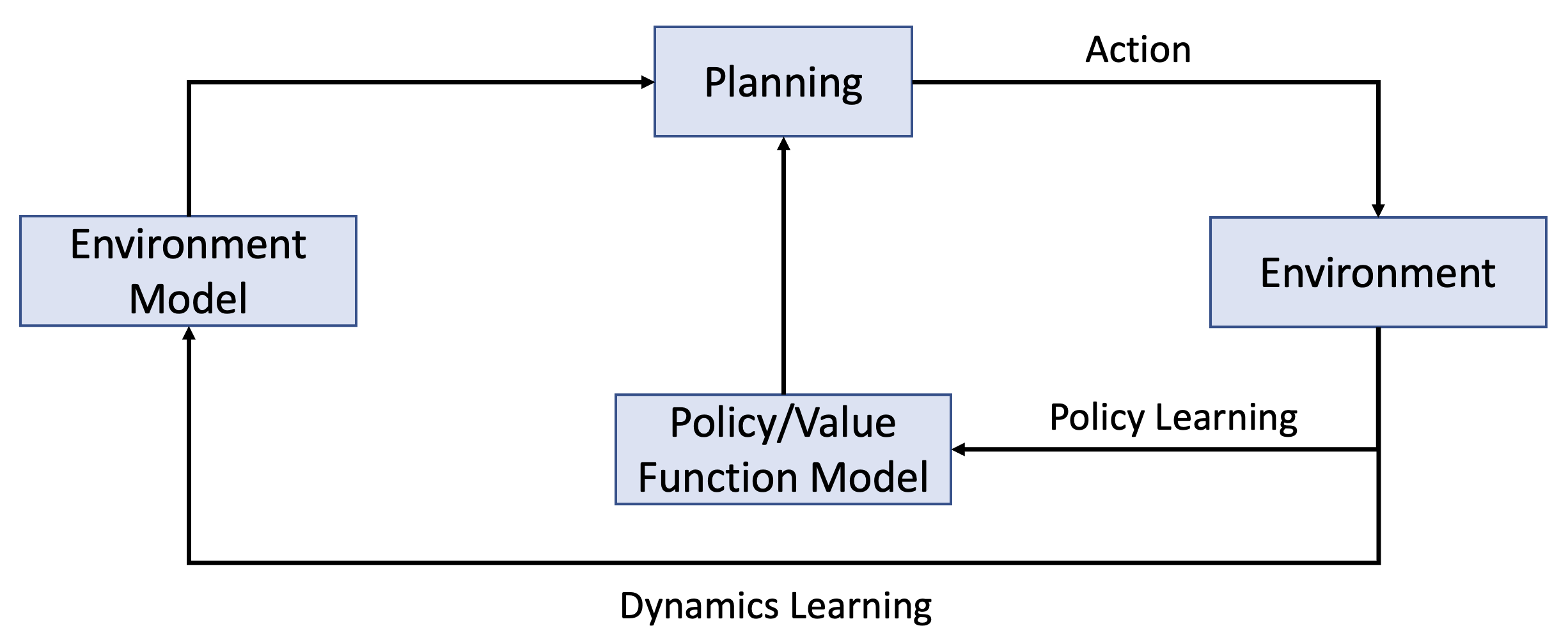}
  \caption{Real-time planning for transportation can be modelled as a sequential decision-making problem. Planning requires making decisions over intractably large state-action spaces in stochastic settings. Learning can aid planning by either 1) learning the dynamics of the environment or 2) directly learning or use information from a policy/value network to improve the planning procedure.}
  \label{fig:learning}
\end{figure}

There are many ways in which we can utilize descriptive and generative models to aid optimization in the transportation domain.
For example, take the high-level description of a sequential decision problem, as shown in Figure~\ref{fig:learning}, in the context of on-demand ride-share.
In on-demand ride-share, at each timestep a new batch of trip requests arrives and must be assigned to vehicles that can service these new requests.
The \textit{planning} component refers to  algorithms or methods responsible for this assignment at each timestep.
In this case, the \textit{environment} refers to the transportation system itself and includes roadways, traffic congestion, weather, trip requests, etc., all of which generate data.
The planner and environment interact sequentially, and as discussed in Section~\ref{subsection:optimization for transportation}, the goal is to select actions that maximize the objective (costs, emissions, and utilization) over time. 

Transportation systems are computationally complex, which makes traditional analytical methods intractable in this setting.
Therefore, we can utilize AI and machine learning to learn condensed representations through data to aid planning.
In Figure~\ref{fig:learning}, we highlight two ways to achieve this.

\subsubsection{Dynamics Learning} Dynamics learning refers to directly learning representations related to the environment itself.
Recall from Section~\ref{subsubsection: real-time control} that the environment consists of the transition function $\mathcal{T}$ and reward function $\mathbf{R}$. Using these two functions, the environment takes as input the current state $s_t$ and action $a_t$ and returns the next state $s_{t+1}$ and reward $r_t$.
The goal of dynamics learning is to learn the transition function $\mathbb{P}[s_{t+1}|s_t, a_t]$ and/or the reward function $\mathbb{P}[r_{t}|s_t, a_t]$ through experience.
%As discussed in Section~\ref{subsubsection: real-time control} this would m
For example, in the context of on-demand ride-share, we could learn a generative model of demand or a traffic congestion model.
These environment models can be used to investigate how certain decisions will play out over time and contextualize decisions in light of future scenarios through \textit{planning}.
Planning in this context can be a greedy algorithm that simply selects the action with the greatest predicted reward using the reward function or use a combination of the transition and reward functions to look-ahead through online search.
A common online search method in this setting is Monte Carlo tree search (MCTS), which we discuss in later sections.
This combination of model-learning and planning is often referred to as model-based RL in the RL literature \cite{moerland2023model}.

\subsubsection{Policy Learning} Alternatively, recent advances in model-free reinforcement learning aim to directly learn policies or value functions that can be embedded directly in the planning component.
This can be done with or without a model of the environment.
For example, recall that the policy $\pi$ is responsible for selecting an action given the current state with the goal of maximizing reward over time.
Therefore a policy $\mathbb{P}[a_t|s_t]$ can be parameterized and learned directly through experience or simulation.
Alternatively, we can learn a value function that predicts the future reward for a state given a policy $v_{\pi}(s)=\mathbb{E}_{\pi}[r_{t+1} + r_{t+2} + \dots | S_t=s]$.
Model-free RL has proven effective in numerous application domains including learning to play video games or other camera-input applications where learning a model of the environment is intractable \cite{arulkumaran2017deep}.
However, as we will discuss, in the transportation domain a model of the environment can often be learned.
In the following sections, we discuss concrete examples of how to combine environment models, policies, and planning to solve unique problems in the transportation domain.

%\subsection{Application Domains}

%There are many ways in which AI can aid optimization in the transportation domain.

%\subsection{Applications}

%\subsubsection{Fixed Line Transit - Samitha, Abhishek}

%- samitha -- line planning

%\subsubsection{On-Demand Transit - Samitha}
%i.e., online VRP

%Multi-modal

%\subsubsection{Energy-Efficient Vehicle Assignment}

%Aron

\begin{copypasted}
\sidenote{copypasted from AAAI-21 paper}
In the U.S.,
28\% of total energy use is for transportation~\cite{eia}. 
While public transit services can be very energy-efficient compared to personal vehicles, their environmental impact is significant nonetheless.
For example, bus transit services in the U.S. may be responsible for up to 21.1 million metric tons of CO$_2$ \cite{ghgemissions} emission every year.

Electric vehicles (EVs) can have much lower operating costs and lower environmental impact during operation than comparable internal combustion engine vehicles (ICEVs), especially in urban areas.
Unfortunately, EVs are also much more expensive than ICEVs: typically, diesel transit buses cost less than \$500K, while electric ones cost more than \$700K, or close to around \$1M with charging infrastructure.
As a result, many public transit agencies can afford only mixed fleets of transit vehicles, which may consist of EVs, hybrid vehicles (HEVs), and ICEVs. 

Public transit agencies that operate such mixed fleets of vehicles face a challenging optimization problem.
First, they need to decide which vehicles are assigned to serving which transit trips.
Since the advantage of EVs over ICEVs varies depending on the route and time of day (e.g.,  advantage of EVs is higher in slower traffic with frequent stops, and lower on highways), the assignment can have a significant impact on energy use and, hence, on costs and environmental impact.  
Second, transit agencies need to schedule when to charge electric vehicles because EVs have limited battery capacity and driving range, and may need to be recharged during the day between serving transit trips. 
Since agencies often have limited charging capabilities (e.g., limited number of charging poles, or limited maximum power to avoid high peak loads on the electric grid), charging constraints can significantly increase the complexity of the assignment and scheduling problem. 

In this chapter, we present a novel problem formulation and algorithms for assigning a mixed fleet of transit vehicles to trips and for scheduling the charging of electric vehicles.
We developed this problem formulation in collaboration with the Chattanooga Area Regional Transportation Authority (CARTA), the public transit agency of Chattanooga, TN, which operates a fleet of EVs, HEVs, and ICEVs.
To solve the problem, we introduce an integer program as well as greedy and simulated annealing algorithms.
Our problem formulation applies to a wide range of public transit agencies that operate fixed-route services.
Our objective is to minimize energy consumption (i.e., fuel and electricity use), which leads to lower operating costs and environmental impact---as demonstrated by our numerical results.
Our problem formulation considers assigning and scheduling for a single day (it may be applied to any number of consecutive days one-by-one), and allows any physically possible re-assignment during the day.
Our formulation also allows capturing additional constraints on charging; for example, CARTA aims to charge only one vehicle at a time to avoid demand charges from the electric utility.
\end{copypasted}

%\subsection{Offline and Online Vehicle Routing Problems}

\begin{copypasted}
\sidenote{copypasted from ACM TOIT paper}
Our study is most closely related to the work of Cauwer et al. and of Wickramanayake and Bandara.
Cauwer et al. \cite{de2017data} use a cascade of ANN and multiple linear regression models as a data-driven energy-consumption prediction method for EVs.
Their study uses vehicle monitoring data for two types of vehicles as time series of tuples with location, vehicle speed, and energy-consumption information, such as battery voltage, current, and SoC. Their dataset also includes road-network data, weather data, and an altitude map. Our approach has some similarity to this study. However, we also use traffic data in our model, which we find to be very helpful for diesel prediction.
Wickramanayake and Bandara \cite{wickramanayake2016fuel} assess three different techniques for predicting the fuel consumption of a long-distance public bus. Their time series of tuples include GPS location, bearing, elevation, distance travelled, speed, acceleration, ignition status, battery voltage, fuel level, and fuel consumption. The authors compare the performance among two ensemble models, random forest and gradient boosting, and one ANN model. However, their study lacks critical parameters, such as road information, traffic, weather, etc. 

Perrotta et al. \cite{perrotta2017application} compare the performance of SVM, RF, and ANN in modeling fuel consumption of a large fleet of trucks. Their features include gross vehicle weight, speed, acceleration, geographical position, torque percentage, revolutions of the engine, activation of cruise control, use of brakes and acceleration pedal, measurement of travelled distance, fuel consumption. The study also integrates some road characteristics.
Based on comparison of the RMSE, MAE, and $R^2$ scores of the prediction, RF gives the best performance. 
Nageshrao et al. \cite{nageshrao2017charging} model the energy consumption of electric buses based on time-dependent factors such as ambient temperature and speed, battery capacity, total mass, battery parameters, etc. They use a NARX based ANN time series predictor to predict the state of charge of the battery.
Gao et al. \cite{gao2018electric} discuss an adaptive wavelet neural network (WNN) based energy prediction. Their study uses features such as day type, temperature, rainfall, the travelled distance, and clarity of the atmosphere. The study groups the trip days based on similar attributes, using Grey Relational Analysis (GRA) and then implements the Adaptive WNN.

There also exist prior efforts to utilize spatial and temporal data to model the energy consumption  of bus transit networks and to estimate costs. Wang et al. \cite{wang2018bcharge} collect GPS records of vehicle position and vehicle status and GPS location details of bus stops and bus transaction data of passenger fares.  Wang et al. \cite{wang2017data} and Hassold et al. \cite{hassold2014improving} obtain historical transaction data. Li et al. \cite{li2019mixed},  Paul et al. \cite{paul2014operation} and Li et al. \cite{li2014transit} gather the distance of each trip in the bus transit network schedule. Santos et al. \cite{santos2016towards}  and  Zhou et al. \cite{zhou2020collaborative} make use of publicly available resources, which provide details of average energy costs per unit of energy, emission rate per unit of energy consumption, and consumption rates per unit distance. In contrast to previous efforts, we collect traffic, elevation and weather data in addition to vehicle position and vehicle status data.
\end{copypasted}

%\subsection{Modern Transit Systems}

\section{Computational Problems and AI Approaches}

In this section, we discuss some key computational problems in which AI has been used in recent work related to transportation.

%\subsection{Classical Line Planning}
%\subsection{Delay Prediction and Headway Management}

\subsection{On-demand Paratransit and Microtransit}

Advances in sensing and smart phone adoption have led to rapid expansion of new demand-responsive modes of transportation in recent years such as ride-share (Uber, Lyft), e-scooters, and bike-share.
In the case of ride-share, users request trips through a smart-phone application and a vehicle provides direct transportation to the user's requested destination.
As users have increasingly utilized demand-responsive services in this way, public transportation agencies are looking for ways to make their own transportation options more demand-responsive and dynamic.
One way in which transit agencies can make their systems more responsive is through microtransit.
Microtransit offers door-to-door service with high-capacity vehicles.
Similarly, paratransit service is a socially beneficial curb-to-curb transportation service provided by public transit agencies for passengers who are unable to use fixed-route transit (e.g. passengers with disabilities).
Paratransit is often required by law and comes with its own unique set of design challenges.

Both paratransit and microtransit are forms of ride-pooling, which can be defined as a direct transportation service that allows trips to be pooled together in a way that users can share rides with other passengers.
By allowing passengers to share trips, ride-pooling has the potential to overcome inefficiencies with traditional ride-share---particularly by reducing vehicle miles travelled (VMT) \cite{sejourne2018price,sivagnanam2022offline}.
However, the design and implementation of ride-pooling systems has numerous challenges.
First, the problem is computationally complex and involves solving complex allocation problems in real-time.
Second, the environment is highly uncertain and dynamic in that traffic, weather, and the distribution of requests can change rapidly over time.

Ride-pooling is a dynamic version of the vehicle routing problem (DVRP) with stochastic trip requests.
In this setting, some customer requests may be known at the time of planning while others are unknown, and some stochastic information may be available about potential future requests.
When a new trip request or batch of trip requests arrive, these requests must be assigned to vehicles that can serve these requests without violating the VRP constraints.
Both microtransit and paratransit include standard constraints, such as that pickups must occur before dropoffs, and each passenger can be served by only one vehicle.
Since paratransit primarily focuses on serving passengers with disabilities, paratransit services operate under the Americans with Disabilities Act (ADA), which enforces time windows as a hard constraint, unlike on-demand microtransit services, thereby requiring strict adherence to such constraints which must be taken into account.

There are many highly efficient myopic solutions available for the ride-pooling problem.
State-of-the-art myopic assignment utilizes shareability graphs to generate a large number of feasible solutions and then batch assigns requests to vehicles through an integer linear program (ILP) \cite{alonso2017demand}.
These methods are highly efficient and can solve the assignment problem to near-optimality at each decision-epoch, but since they do not take into account future information, they may not be optimal when evaluated over the course of a full day.
However, the availability of stochastic information and environment models presents an opportunity to utilize AI to learn more non-myopic solutions that can make decisions in the context of expected future demand and environment conditions.
There are two primary methods for designing non-myopic solutions to the DVRP.
First, we can utilize historical trip requests and demand to build a simulator that can be used to train a value function that encodes the value of a future state in the context of expected reward over the entire day \cite{joe2020deep}.
This approach can be categorized as model-free reinforcement learning and is computationally efficient at run-time.
However, it requires a large amount of high-quality data for the simulator and can perform poorly when real-time conditions deviate greatly from historical patterns.
The second approach is to utilize dynamics learning to create a generative demand model that can be used with online search to evaluate current actions in the context of future scenarios \cite{wilbur2022iccps}.
This method is more adaptive to changing environment dynamics at run-time since traffic and demand models can be updated based on current conditions.
However, this approach is less scalable since it requires computationally intensive online search to run at each decision epoch.

%The vehicle routing problem (VRP) is a well-known combinatorial optimization problem that seeks to assign a fleet of vehicles to routes to serve a set of customers/requests. 
%Many real-world use cases of transportation agencies are modeled by the dynamic version of the problem (DVRP) with stochastic trip requests include ride-share, ride-pooling, microtransit and paratransit.
%In such settings, some customer requests may be known at the time of planning while others are unknown, and some stochastic information may be available about potential future requests~\cite{bent2004scenario}. 

%Although the dynamic and the stochastic versions have traditionally been tackled separately, various works showed that dynamic planning could use the stochastic information to improve performance \cite{hvattum2006solving, bent2004scenario}. 
%There are three broad approaches for solving DVRPs with stochastic requests. 
%First, as requests arrive, a group of requests can be batched together, and routes can be optimized myopically for the particular batch in an \textit{online} manner~\cite{alonso2017demand}. 
%Second, the routing problem can be solved to maximize a non-myopic utility function by learning a policy in an \textit{offline} manner that maps any given state of the problem to an action (i.e. a route plan for the vehicles)~\cite{nazari2018reinforcement}. 
%Third, a combination of offline computation and online heuristics can be used for non-myopic planning~\cite{bent2004scenario,joe2020deep}.

\subsection{Prediction of Energy Usage}
\label{subsec:energy_prediction}

Transportation accounts for 28\% of total energy consumption in the U.S.\cite{eia}, leading to significant environmental consequences such as urban air pollution, greenhouse gas emissions. Shifting from personal vehicles to public transit systems offers a substantial opportunity to reduce energy usage and environmental impact. However, public transit itself consumes substantial energy, with U.S. bus services alone emitting around 19.7 million metric tons of CO2 annually~\cite{EPA-420-F-19-047}. 
The adoption of electric vehicles (EVs) can help decrease the environmental impact, but their higher costs result in a combination of EVs, hybrids, and internal combustion engine vehicles (ICEVs) in transit agency fleets. This creates a challenging optimization problem for transit agencies: determining which vehicles should be assigned to specific transit trips. Since the advantages of EVs over ICEVs vary depending on factors like route and time of day (e.g. EVs are more beneficial in slower traffic with frequent stops and less advantageous on highways), the assignment of vehicles can significantly affect energy use and, consequently, environmental impact. The heart of operational optimization of a transit agency lies in the challenge of precisely forecasting the electricity and fuel consumption of transit vehicles. Accurate prediction of electricity and fuel consumption for transit vehicles is difficult due to various factors like vehicle type, traffic conditions, and road characteristics. Leveraging sensor-based technologies, data analytics, and machine learning can address these challenges, allowing the development of a comprehensive framework for route-level energy prediction in public transit.

Within this subsection, we introduce an innovative framework designed for data-driven offline energy consumption prediction at the route level~\cite{ayman2020data,ayman2022data}. Our framework is specifically tailored for mixed-vehicle transit fleets, considering diesel (ICEV), hybrid and electric vehicles to accommodate the operational diversity of the fleet. To evaluate its efficacy, we utilize real-world data obtained from the bus fleet of a public transit authority from a midsize city in U.S.

% We also present a framework and novel algorithms for cleaning and integrating time series data from multiple sources into sets of samples with fixed-dimensional feature space\begin{extension}, including a machine-learning based approach for accurately mapping noisy locations to road segments. 
% We use this dataset to train machine-learning models for energy-use prediction (deep neural networks, linear regression, and decisions trees) and study their performance, focusing on the impact of including or omitting certain data sources.

\subsubsection{Data Processing Framework}
Before applying machine-learning models for energy usage prediction, we must process the time series data recorded from the vehicles by cleaning it, generating samples with a fixed-dimension feature space, and incorporating data from other sources, including traffic and weather data. For each bus, the recorded data is a series of datapoints, numbered $i = 0, 1, 2 \ldots$, where each datapoint is a tuple of a timestamp $TS_i$, a location $L_i$, etc.
For EVs, each datapoint $i$ includes a battery current $A_i$, a battery voltage $V_i$, a battery state of charge (\%), and a charging cable status (0 or 1). For diesel and hybrid vehicles, instead of battery data, datapoint contains fuel usage in gallons. First, we filter out specific data points in our analysis to achieve our objective of predicting energy usage. First, we exclude data recorded when a bus is either waiting in the garage or undergoing charging. We also remove data points where the charging cable status indicates active charging (for electric vehicles).

In order to create labels for our supervised energy prediction model, we estimate the energy usage from the recorded vehicle data.
For diesel buses, we can determine the amount of fuel consumed between two consecutive data points by calculating the change in the total fuel used. For electric buses, estimating energy usage is more intricate. The state of charge values, recorded with limited precision, can be used to approximate energy usage. However, for accurate values, we rely on estimates derived from recorded battery current ($A$) and voltage ($V$) values. Instantaneous power usage (in Watts) at any given time can be computed as the product of $A$ and $V$. We estimate the energy used (in Joules), denoted as $E_i$, between consecutive data points ($i-1$) and ($i$) using $E_{i} = A_{i} \cdot V_{i} \cdot \left( TS_i - TS_{i-1}\right)$.
Here, $TS_i$  represents the timestamp of data point $i$ in seconds. This formula provides highly accurate estimates since current and voltage values are recorded at least once per second. To ensure the validity of our estimates, we conducted comparisons against changes in $SoC$ across numerous data points, confirming their unbiased nature.

One of the important steps in the data processing framework is \emph{cleaning and mapping the GPS location to roads}. The recorded vehicle locations, based on GPS, inherently contain noise. Some locations may fall on streets or parking lots where buses cannot traverse, posing challenges for accurate distance calculations and integration with other data sources. To mitigate this noise, we combine the recorded vehicle locations ($L_i$) with a street-level map obtained from OpenStreetMap (OSM) \cite{haklay2008openstreetmap}. OSM represents roads using disjoint segments known as OSM features, each with unique identifiers and properties. To align the vehicle locations with the street-level map, we employ two novel methods: a heuristic algorithm and a machine-learning approach. 
For mapping recorded locations ($L_i$) to OSM segments (road segments), we consider nearby OSM segments based on geographical distance. We create an R-tree spatial index for the street-level map and intersect it with a bounding disk around $L_i$. This intersection yields a set of potential road segments, which we filter based on road types suitable for buses. 
For heuristic approach, we count the number of preceding and following locations near each nearby segment and select the segment with the most nearby locations. For machine-learning approach, rather than directly outputting the correct segment from the nearby candidate segments, we estimate their likelihood using a regression model. The model incorporates variables such as distance between the location and the candidate OSM segment, road type, and distances to the set of preceding and following locations. The model outputs a value between 0 and 1, indicating the likelihood of correct mapping. We apply the regression model to each nearby candidate segment and select the one with the highest likelihood for mapping the location.

\pgfplotstableread[col sep=comma]{energy_prediction/data/ML_Final_Comparison_Results_All.csv}\noise
\pgfplotstableread[col sep=comma]{energy_prediction/data/Noise_Results_Heuristic_All.csv}\HeuristicNoise
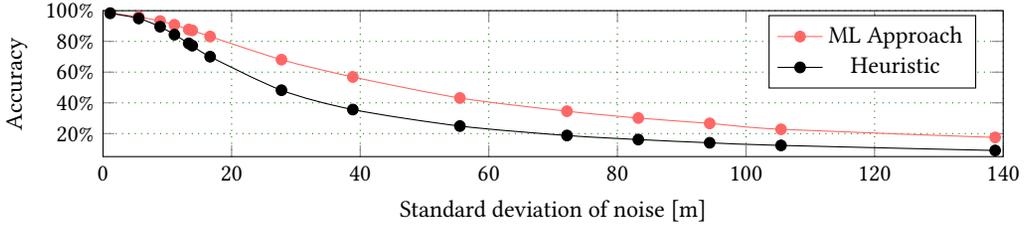
\begin{figure}[t]
    \centering
    \begin{tikzpicture}[font=\small]
        \begin{axis}
        [
            ylabel=Accuracy,
            xlabel={Standard deviation of noise [m]},
            ylabel style = {align = center},
            	grid=major, 
            yticklabel=\pgfmathprintnumber{\tick}$\%$,
            xticklabel=\pgfmathprintnumber{\tick},
            legend pos=north east,
            label style = {align = center},
            width=0.975\linewidth,
            height=3.5cm,
            xmin = 0,xmax = 140,
            ymin = 5, ymax = 100
        ]
            \addplot [red!60, smooth, mark=*] table [x=NoiseDistance, y=Accuracy] {\noise};
            \addlegendentry{ML Approach}
            \addplot [black, smooth, mark=*] table [x=NoiseDistance, y=Accuracy] {\HeuristicNoise};
            \addlegendentry{Heuristic}
        \end{axis}
    \end{tikzpicture}
    \caption{Accuracy of mapping noisy locations to road segments using heuristic and machine-learning approach. Accuracy is measured as the fraction of locations that are mapped to the correct road segment.}
    \label{fig:algorithm_results}
    \vspace{-1em}
\end{figure}

For evaluating the accuracy of different approaches for mapping noisy locations to OSM features, we first generate routes using a street-level map, selecting locations precisely on the roads, and then introduce random noise to these locations (noise is generated from a two-dimensional Gaussian distribution with zero mean)\footnote{The synthetic location traces are solely used for evaluating the mapping approaches, as we require ground-truth segments. However, for training and evaluating energy-use prediction, we utilize real location traces obtained from vehicles.}. The standard deviation of the noise is varied from 1 meter to 140 meters in both directions. Finally, we map the noisy locations to road segments using both the heuristic and machine-learning approaches with decision tree regression. Accuracy is measured as the ratio of correctly mapped locations.
In \cref{fig:algorithm_results}, the accuracy of both approaches is compared across different levels of noise. At the lowest noise level (1.1 meters), both the heuristic algorithm and the machine-learning approach achieve accuracy above 98\%. As expected, the accuracy of both approaches decreases as the noise level increases. However, the machine-learning approach outperforms the heuristic algorithm at higher noise levels. 

Once we map the location traces to road segments, we can generate samples by dividing the time series data into segments based on the traveled road segments. Each sample represents a maximal continuous travel on a specific road segment and includes the starting and end locations, starting and end times, and the sum of energy used during the travel. To accurately calculate the distance traveled for each sample, we obtain the geometry of the corresponding road segment from OSM and identify the line segments that the bus actually traveled. The distance traveled for each sample is calculated based on the partial distance on the starting line segment, the complete distance of the intermediate line segments, and the partial distance on the ending line segment.

Finally, we incorporate additional features for the prediction model. These features encompass elevation changes within the samples, weather-related attributes like temperature, humidity, visibility, precipitation, and wind speed, as well as traffic data such as the speed ratio between actual speed and free-flow speed.

\subsubsection{Prediction Models}
For prediction of energy usage, we apply three different machine-learning models: artificial neural network, linear regression, and decision tree regression.
The input of the energy prediction models (i.e. training or test set) is a set of samples, where each sample is a tuple of distance travelled, various road-type features, elevation change, various weather features, various traffic features, and energy used as the target feature.
Before training and testing, we map categorical variables (e.g. road type) into sets of binary features using one-hot encoding.
We train all three models to minimize \emph{mean squared error}~(MSE).

For our neural-network based models, we use feed-forward deep neural networks with one input layer, multiple hidden layers, and one output layer. Sigmoid activation is used in the hidden layers, and linear activation is used in the output layer for both diesel and electric. The models are optimized with the \emph{Adam} optimizer~\cite{kingma2014adam} and a learning rate of $0.001$, implemented using Keras~\cite{chollet2015keras}, a high-level API of TensorFlow.
We use standard multiple linear regression and decision tree regression~\cite{dtr} as other prediction models.
For implementing the models, we use the implementation provided by the \emph{scikit-learn} Python library. Decision tree regression builds a tree structure based on the training samples, where each node represents a decision based on the value of a feature variable, and leaf nodes provide predictions.
We opted for neural networks due to their exceptional predictive capabilities, a fact supported by our numerical findings. In comparison, linear and decision tree regression techniques do not exhibit the same level of performance, although their outcomes are more straightforward to comprehend and articulate. For instance, linear regression highlights the direct relationship between input variables and target features, providing a clear understanding of their influence.

\subsubsection{Numerical Evaluation}

\paragraph{Comparison of Prediction Models for Short trips}

We assess the performance of three machine-learning models in predicting energy usage for short trip segments. These segments pose a challenge due to their limited distance and duration. \cref{fig:short_trip_comparison} displays the results in terms of mean squared error (MSE) for all three models for both diesel and electric vehicles. The artificial neural network (ANN) demonstrates superior performance compared to the other models, both for electric and diesel vehicles.
\pgfplotstableread[col sep=comma]{energy_prediction/data/Gillig2014_Shorter_Interval_Loss_PreCovid.csv}\Diesel
\pgfplotstableread[col sep=comma]{energy_prediction/data/BYD2017_Shorter_Interval_Loss_PreCovid.csv}\Electric
\begin{figure}[t]
\begin{subfigure}[b]{0.48\linewidth}
    \centering
    \begin{tikzpicture}[font=\small]
    \begin{axis}
         [
        ybar,
        width=0.9\linewidth,
        ymin=0,
        xtick=data,
        xticklabel style={text width=2.5cm, align=center},
        xtick pos=left,
       ylabel=\emph{MSE},
        ylabel style = {align = center},
            grid=major,
        xtick = data,
        xticklabels from table={\Electric}{Models},
        yticklabel=\pgfmathprintnumber{\tick},
        tick label style={font=\small}, 
        legend pos=north east,
        height=3cm,
        ymin=0.0075,
        legend style={at={(1,1)},anchor=north east,font=\small}
        ]
        \addplot [draw=red,fill=red!40] table [x expr=\coordindex, y=MSE] {\Electric};
    \end{axis}
\end{tikzpicture}
\label{Electric_MSE}
    \caption{\emph{Electric}}
\end{subfigure}
\begin{subfigure}[b]{0.48\linewidth}
\begin{tikzpicture}[font=\small]
    \begin{axis}
        [
        ybar,
        width=0.9\linewidth,
        ymin=0,
        xtick=data,
        xticklabel style={text width=2.5cm, align=center},
        xtick pos=left,,
	       ylabel=\emph{MSE},
            ylabel style = {align = center},
            	grid=major,
            xtick = data,
            xticklabels from table={\Diesel}{Models},
            yticklabel=\pgfmathprintnumber{\tick},
            tick label style={font=\small}, 
            legend pos=north east,
            height=3cm,
            ymin=0.0004,
            legend style={at={(1,1)},anchor=north east,font=\small}
        ]
            \addplot [draw=blue,fill=blue!40] table [x expr=\coordindex, y=MSE] {\Diesel};
        \end{axis}
\end{tikzpicture}
    \caption{\emph{Diesel}}
    \label{fig:Diesel_MSE}
\end{subfigure}
\caption{{Comparison of different energy prediction models based on \emph{mean squared error} (MSE) for electric and diesel vehicles.}}
\label{fig:short_trip_comparison}
\vspace{-1.5em}
\end{figure}
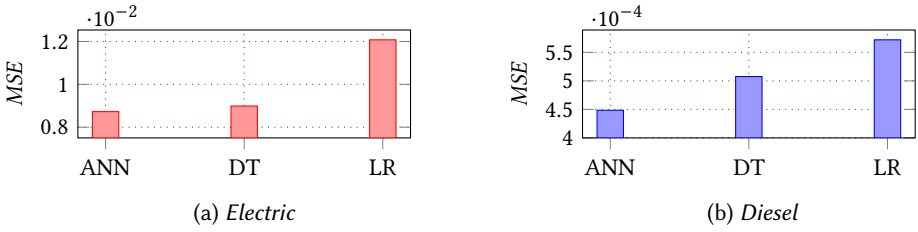

\paragraph{Comparison of Prediction Models for Longer Trips}
We also examine the performance of our models in predicting energy usage for longer trips. To accomplish this, we segment our time series into extended durations, ranging from 10 minutes to 6 hours. For each trip duration, we generate a collection of examples and utilize our models to predict energy consumption for each example. Subsequently, we compare the sum of these predictions with the actual energy usage recorded for the entire trip.
\cref{fig:long_trip_comparison} shows the relative prediction error for trips of various lengths. When analyzing various trip lengths, we observe that the average error values, computed across multiple trips, tend to be lower for longer trips. This is an expected outcome because with larger sample sizes, the individual errors of numerous samples cancel out each other when using an unbiased prediction model. In the case of diesel vehicles, we discover that the artificial neural network (ANN) consistently outperforms the other models across all trip lengths, exhibiting significant improvement. However, for electric vehicles, both the ANN and decision tree (DT) models perform equally well for the majority of trip lengths.

 \pgfplotstableread[col sep=comma]{energy_prediction/data/Gillig2014_Longer_Interval_Loss_PreCovid.csv}\Diesel
\pgfplotstableread[col sep=comma]{energy_prediction/data/BYD2017_Longer_Interval_Loss_PreCovid_extra.csv}\Electric
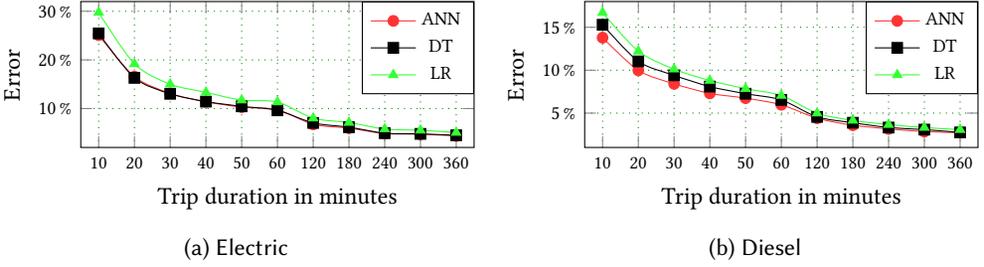
\begin{figure}[h!]%
\begin{subfigure}[b]{0.49\linewidth}
    \centering
    \begin{tikzpicture}[font=\small]
    \begin{axis}
        [
            xlabel=Trip duration in minutes,
	       ylabel=Error,
            ylabel style = {align = center},
            	grid=major,
            xtick = data,
            xticklabels=
            {10,20,30,40,50,60,120,180,240,300,360},
            yticklabel=\pgfmathprintnumber{\tick}\,$\%$,
            tick label style={font=\scriptsize}, 
            legend pos=north east,
            width=\linewidth,
            height=3.5cm,
            xmin = -0.5,xmax = 10.5,
            ymin = 2, ymax = 32,
            legend style={at={(1,1)},anchor=north east,font=\scriptsize}
        ]
            \addplot [red!80, smooth, mark=*] table [x expr=\coordindex, y=FF] {\Electric};
            \addlegendentry{ANN}
            \addplot [black, smooth, mark=square*] table [x expr=\coordindex, y=DT] {\Electric};
            \addlegendentry{DT}
            \addplot [green!80, smooth, mark=triangle*] table [x expr=\coordindex, y=LR] {\Electric};
            \addlegendentry{LR}
    \end{axis}
\end{tikzpicture}
    \caption{{Electric}}
    \label{fig:Electric_LongerTrips}
\end{subfigure}
\begin{subfigure}[b]{0.49\linewidth}
\begin{tikzpicture}[font=\small]
    \begin{axis}
        [
            xlabel=Trip duration in minutes,
	       ylabel=Error,
            ylabel style = {align = center},
            	grid=major,
            xtick = data,
            xticklabels=
            {10,20,30,40,50,60,120,180,240,300,360},
            yticklabel=\pgfmathprintnumber{\tick}\,$\%$,
             tick label style={font=\scriptsize}, 
            legend pos=north east,
            width=\linewidth,
            height=3.5cm,
            xmin = -0.5,xmax = 10.5,
            ymin = 1, ymax = 18,
            legend style={at={(1,1)},anchor=north east,font=\scriptsize}
        ]
            \addplot [red!80, smooth, mark=*] table [x expr=\coordindex, y=FF] {\Diesel};
            \addlegendentry{ANN}
            \addplot [black, smooth, mark=square*] table [x expr=\coordindex, y=DT] {\Diesel};
            \addlegendentry{DT}
            \addplot [green!80, smooth, mark=triangle*] table [x expr=\coordindex, y=LR] {\Diesel};
            \addlegendentry{LR}
        \end{axis}
\end{tikzpicture}

    \caption{{Diesel}}
    \label{fig:Diesel_LongerTrips}
\end{subfigure}
\caption{{Energy prediction error for longer trips, consisting of many samples, with neural network (ANN), decision tree (DT), and linear regression~(LR).}}
\label{fig:long_trip_comparison}
\vspace{-1em}
\end{figure}

\subsection{Energy-Efficient Vehicle Assignment}

\begin{copypasted}
\sidenote{copypasted from AAAI-21 paper}

\subsubsection{Problem Formulation}
\newcommand{\MT}[0]{\ensuremath{T}}
\newcommand{\DMT}[2]{\ensuremath{D({#1}, {#2})}}

\paragraph{Vehicles}
We consider a transit agency that operates a set of \emph{buses}~$\mathcal{V}$.
Note that we will use the terms \emph{bus} and \emph{vehicle} interchangeably.
Each bus $v \in \mathcal{V}$ belongs to a \emph{vehicle model}~$M_v \in \mathcal{M}$, where $\mathcal{M}$ is the set of all vehicle models in operation.
We divide the set of vehicle models $\mathcal{M}$ into two disjoint subsets: liquid-fuel models $\mathcal{M}^\text{gas}$ (e.g., diesel, hybrid), and electric models $\mathcal{M}^\text{elec}$.
Based on discussions with CARTA, we assume that vehicles belonging to a liquid-fuel model can operate all day without refueling.
On the other hand, vehicles belonging to an electric model have limited battery capacity, which might not be enough for a whole day.
For each electric vehicle model $m \in \mathcal{M}^\text{elec}$, 
we let~$C_m$ denote the \emph{battery capacity} of vehicles of model~$m$.

\paragraph{Locations} 
Locations $\mathcal{L}$ include bus stops, garages, and charging stations in the transit network.

\paragraph{Trips}
During the day, the agency has to serve a given set of \emph{transit trips}~$\mathcal{T}$ using its buses. 
Based on discussions with CARTA, we assume that locations and time schedules are fixed for every trip. 
A bus serving trip $t \in \mathcal{T}$ leaves  from trip origin $t^{\text{origin}} \in \mathcal{L}$ at time $t^{\text{start}}$ and arrives at destination $t^{\text{destination}} \in \mathcal{L}$ at time $t^{\text{end}}$. 
Between $t^{\text{origin}}$ and $t^{\text{destination}}$, the bus must pass through a series of stops at fixed times; however, since we cannot re-assign a bus during a transit trip, the locations and times of these stops are inconsequential to our model.
Finally, we assume that any bus may serve any trip. Note that it would be straightforward to extend our model and algorithms to consider constraints on which buses may serve a trip (e.g., based on passenger capacity).

\paragraph{Charging}
To charge its electric buses,
the agency operates a set of \emph{charging poles} $\mathcal{CP}$, which are typically located at bus garages or charging stations in practice.
We let $cp^\text{location} \in \calL$ denote the location of charging pole $cp \in \mathcal{CP}$.

For the sake of computational tractability, we use a discrete-time model to schedule charging, which divides time into uniform-length \emph{time slots} $\mathcal{S}$. 
A time slot $s \in \mathcal{S}$ begins at~$s^{\text{start}}$ and ends at~$s^{\text{end}}$.
A charging pole~$cp \in \mathcal{CP}$ can charge $P(cp, M_v)$ energy to one electric bus $v$ in one time slot.
We will refer to the combination of a charging pole $cp \in \mathcal{CP}$ and a time slot $s \in \mathcal{S}$ as a \emph{charging slot} $(cp, s)$;
and we let $\mathcal{C} = \mathcal{CP} \times \mathcal{S}$ denote the set of charging slots.

\paragraph{Non-Service Trips}
Besides serving transit trips, buses may also need to drive between trips or charging poles.
For example, if a bus has to serve a trip that starts from a location that is different from the destination of the previous trip, the bus first needs to drive to the origin of the subsequent trip.
An electric bus may also need to drive to a charging pole after serving a transit trip to recharge, and then drive from the pole to the origin of the next transit trip.
We will refer to these deadhead trips, which are driven outside of revenue service, as \emph{non-service trips}.
We let $\MT(l_1, l_2)$ denote the non-service trip from location $l_1 \in \mathcal{L}$ to $l_2 \in \mathcal{L}$;
and we let $\DMT{l_1}{l_2}$ denote the time duration of this non-service trip.

\subsubsection{Solution Space}
Our primary goal is to assign a bus to each transit trip. 
Additionally, electric buses may also need to be assigned to charging slots to prevent them from running out of power.

\paragraph{Solution Representation}
We represent a solution as a \emph{set of assignments} $\calA$. 
For each trip $t \in \mathcal{T}$, a solution assigns exactly one bus $v \in \mathcal{V}$ to serve trip $t$; this assignment is represented by the relation $\langle v,t \rangle \in \calA$.
Secondly, each electric bus $v$ must be charged before its battery state of charge drops below the safe level for operation.
A solution assigns at most one electric bus $v$ to each charging slot $(cp, s) \in \mathcal{C}$; this assignment is represented by the relation $\langle v, (cp, s) \rangle \in \calA$.
We assume that when a bus is assigned for charging, it remains at the charging pole for the entire duration of the corresponding time slot.

\paragraph{Constraints}
If a bus $v$ is assigned to serve an earlier transit trip $t_1$ and a later trip $t_2$, then the duration of the non-service trip from $t_1^\text{destination}$ to $t_2^\text{origin}$ must be less than or equal to the time between $t_1^\text{end}$ and $t_2^\text{start}$.
Otherwise, it would not be possible to serve $t_2$ on time.
We formulate this constraint~as:
\begin{align}
\forall t_1, t_2 \in \calT; ~ t_1^\text{start} \leq t_2^\text{start}; ~ \langle v,t_1 \rangle \in \calA ;  ~ \langle v,t_2 \rangle \in \calA : \nonumber \\
t_1^{\text{end}} + \DMT{t_1^{\text{destination}}}{t_2^{\text{origin}}} \leq t_2^{\text{start}}
\label{equ:constraint_1a}
\end{align}
Note that if the constraint is satisfied by every pair of consecutive trips assigned to a bus, then it is also satisfied by every pair of non-consecutive trips assigned to the bus.

We need to formulate similar constraints for non-service trips to, from, and between charging slots:
\begin{align}
\forall t \in \calT; & ~ (cp, s) \in \calC; \, t^\text{start} \leq s^\text{start}; \,  \langle v,t \rangle, \langle v, (cp,s) \rangle \in \calA : \nonumber \\
& t^{\text{end}} +\DMT{t^{\text{destination}}}{cp^{\text{location}}} \leq s^{\text{start}}
\label{equ:constraint_1b} \\
\forall t \in \calT; & ~ (cp, s) \in \calC; t^\text{start} \geq s^\text{start};  \langle v,t \rangle, \langle v, (cp,s) \rangle \in \calA : \nonumber \\
& s^{\text{end}} +\DMT{{cp^{\text{location}}}}{{t^{\text{origin}}}} \leq t^{\text{start}}
\label{equ:constraint_1c} \\
\forall (cp_1, s_1), & ~ (cp_2, s_2) \in \calC;  s_1^\text{start} \leq s_2^\text{start};   \langle v,(cp_1, s_1) \rangle, \langle v, (cp_2,s_2) \rangle \in \calA : \nonumber \\
& s_1^{\text{end}} + \DMT{cp_1^{\text{location}}}{cp_2^{\text{location}}} \leq s_2^{\text{start}}
\label{equ:constraint_1d}
\end{align}

We also need to ensure that electric buses never run out of power.
First, we let $\calN(\calA, v, s)$ denote the set of all non-service trips that bus $v$ needs to complete by the end of time slot~$s$ according to the set of assignments $\calA$.
In other words, $\calN(\calA, v, s)$ is the set of all necessary non-service trips to the origins of transit trips that start by $s^\text{end}$ and to the locations of charging slots that start by $s^\text{end}$.
Next, we let $E(v, t)$ denote the amount of energy used by bus $v$ to drive a transit or non-service trip~$t$.
Then, we let $e(\calA, v, s)$ be the amount of energy used by bus~$v$ for all trips completed by the end of time slot $s$: 
\begin{equation}
e(\calA, v, s) = \sum_{t \in \calN(\calA, v, s)} E(v, t) + \quad \quad \sum_{\mathclap{t \in \calT, \, \langle v, t \rangle \in \calA , \, t^\text{end} \leq s^\text{end}}}  \quad ~~ E(v, t)    
\end{equation}
Similarly, we let $r(\calA, v, s)$ be the amount of energy charged to bus $v$ by the end of time slot $s$:
\begin{equation}
r(\calA, v, s) = \sum_{(cp, \hat{s}) \in \calC, \, \langle v, (cp,\hat{s}) \rangle \in \calA, \, \hat{s}^\text{end} \leq s^\text{end} } P(cp, M_v)
\end{equation}
Since a bus can be assigned for charging only to complete time slots, the minima and maxima of its battery level will be reached at the end of time slots.
Therefore, we can express the constraint that the battery level of bus $v$ must always remain between $0$ and the battery capacity $C_{M_v}$ as
\begin{equation}
    \forall v \in \calV, \forall s \in \calS: ~ 0 < r(\calA, v, s) - e(\calA, v, s) \leq C_{M_v} .
    \label{eq:energy_constr}
\end{equation}
Note that we can give vehicles an initial battery charge by adding ``virtual'' charging slots before the day starts.

\subsubsection{Objective}
Our objective is to minimize the energy use of the transit vehicles.
We can use this objective to minimize both environmental impact and operating costs by imposing the appropriate cost factors on the energy use of liquid-fuel and electric vehicles. 
We let $K^\text{gas}$ and $K^\text{elec}$ denote the unit costs of energy use for liquid-fuel and electric vehicles, respectively.
Then, by applying the earlier notation $e(\calA, v, s)$ to all vehicles, we can express our objective as 
\begin{equation}
\min_{\calA} ~~\quad \sum_{\mathclap{v \in \calV: \, M_v \in \calM^\text{gas}}} ~ K^\text{gas} \! \cdot e(\calA, v, s_\infty) + ~~\quad \sum_{\mathclap{v \in \calV: \, M_v \in \calM^\text{elec}}} ~ K^\text{elec} \! \cdot e(\calA, v, s_\infty)
\label{equ:obj_main}
\end{equation}
where $s_\infty$ denotes the last time slot of the day.

\subsubsection{Algorithms}
Since this optimization problem is NP-hard, we first present an integer program to find optimal solutions for smaller instances.
Then, we introduce efficient greedy and simulated annealing algorithms,
which scale well for larger instances. 
\sidenote{need to revise}
Due to lack of space, we include less important subroutines in~\cite{sivagnanam2021minimizing}.

\subsubsection{Integer Program}
\label{sec:ip}

\paragraph{Variables}
Our integer program has five sets of variables.
Three of them are binary to indicate assignments and non-service trips.
First, $a_{v,t} = 1$ (or $0$) indicates that trip $t$ is assigned to bus $v$ (or that it is not). 
Second, $a_{v, (cp,s)} = 1$ (or $0$) indicates that charging slot $(cp,s)$ is assigned to electric bus $v$ (or not). 
Third, $m_{v,x_1,x_2} = 1$ (or $0$) indicates that bus $v$ takes the non-service trip between a pair  $x_1$ and $x_2$ of transit trips and/or charging slots (or not). 
Note that for requiring non-service trips (see \cref{equ:constraint_1a,equ:constraint_1b,equ:constraint_1c,equ:constraint_1d}), we will treat transit trips and charging slots similarly since they induce analogous constraints.
There are also two sets of continuous variables.
First, $c^v_s \in [0, C_{M_v}]$ represents the amount of energy charged to electric bus $v$ in time slot $s$. 
Second, $e^v_s \in [0, C_{M_v}]$ represents the battery level of electric bus $v$ at the start of time slot~$s$ (considering energy use only for trips that have ended by that time).
Due to the continuous variables, our program is a mixed-integer program.

\paragraph{Constraints}
First, we ensure that every transit trip is served by exactly one bus:
\begin{equation*}
\forall t \in \calT: ~ \sum_{v \in \mathcal{V}} a_{v,t} = 1
\end{equation*}
Second, we ensure that each charging slot is assigned at most one electric vehicle: 
\begin{equation*}
    \forall (cp,s) \in \calC: \sum_{ \forall v \in \calV: ~  M_v \in \mathcal{M}^{\text{elec}}} a_{v,(cp,s)} \leq 1
\end{equation*}

Next, we ensure that \cref{equ:constraint_1a,equ:constraint_1b,equ:constraint_1c,equ:constraint_1d} are satisfied.
We let $F(x_1, x_2)$ be \emph{true} if a pair $x_1, x_2$ of transit trips and/or charging slots satisfies the applicable one from \cref{equ:constraint_1a,equ:constraint_1b,equ:constraint_1c,equ:constraint_1d}; and  let it be \emph{false} otherwise.
Then, we can express these constraints as follows:
\begin{equation*}
\forall v \in \calV, \forall x_{1}, x_{2}, \lnot F(x_1, x_2) : ~ a_{v,x_1} + a_{v, x_2} \leq 1
\end{equation*}

When a bus $v$ is assigned to both $x_1$ and $x_2$, but it is not assigned to any other transit trips or charging slots in between (i.e., if $x_1$ and $x_2$ are consecutive assignments), then bus~$v$ needs to take a non-service trip: 
\begin{equation*}
m_{v, x_1, x_2} \geq a_{v, x_1} + a_{v, x_2} - 1
    -  {\sum_{\substack{x \in \calT \,\cup\, \calC:~ x_1^\text{start} \leq x^\text{start} \leq x_2^\text{start}}}}  a_{v,x}
\end{equation*}
Note that if $x_1$ ends at the same location where $x_2$ starts, then the non-service trip will take zero time and energy.

Finally, we ensure that the battery levels of electric buses remain between zero and capacity.
First, for each slot $s$ and electric bus $v$, the amount of energy charged $c^v_s$ is subject to
\begin{equation*}
    c^v_{s} \leq \sum_{(cp,s) \in \calC} a_{v, (cp,s)} \cdot P(cp, M_v) .
\end{equation*}
Then, for the $(n+1)$\textit{th} time slot $s_{n+1}$ and for an electric bus $v$, we  can  express variable $e_{s_{n+1}}^v$~as 
\begin{align*}
e^v_{s_{n+1}} \!\!= e^v_{s_n} \!\! + 
       c^v_{s_n} \!\!&- ~~~~~~~~~ {\sum_{\mathclap{\substack{t \in \calT : \, s_n^\text{start} < t^\text{end} \leq s_n^\text{end}}}}} ~~~~~~~~~ a_{v,t} \! \cdot \! E(v,t) \\
     &- ~~~~~~~~~ {\sum_{ \mathclap{\substack{ x_1, x_2 : \, s_n^\text{start} < x_2^\text{start} \leq s_n^\text{end} }}}} ~~~~~~~~~ m_{v, x_1, x_2} \! \cdot \! E(v, \MT(x_1, x_2))
\end{align*}
where $s_{n}$ is the $(n)$\textit{th}  slot. Note that since $e^v_s \in [0, C_{M_v}]$, this constraint ensures that \cref{eq:energy_constr} is satisfied.

\paragraph{Objective}
We can express \cref{equ:obj_main} as minimizing
\begin{align*}
    \smashoperator{\sum_{v \in \calV}} K^{M_v} \! \left[ \sum_{t \in \calT} a_{v,t} \!\cdot\! E(v, t) + \smashoperator{\sum_{x_1, x_2 \in \calT \cup \calC}} m_{v,x_1,x_2} \!\cdot\! E(v, \MT(x_1, x_2)) \right] 
\end{align*}
where $K^{M_v}$ is $K^\text{elec}$ if $M_v \in \calM^\text{elec}$ and $K^\text{gas}$ otherwise.

%\paragraph{Complexity}
%The integer program contains both variables and constraints in the order of $\mathcal{O}(|\mathcal{V}|\cdot|\mathcal{T}|^2)$.

\subsubsection{Greedy Algorithm}
\label{sec:greedy}

Next, we introduce a polynomial-time greedy algorithm.
The key idea of this algorithm is to choose between assignments based on a \emph{biased cost} instead of the actual cost.

\begin{algorithm}[!h]
 \caption{$\textbf{BiasedCost}(\mathcal{A}, v, x, \alpha)$}
 \label{algo:energy_cost}

\uIf {$x \in \calT$}
{
 $\textit{cost} \leftarrow E(v,x)$
}
\Else
{
 $\textit{cost} \leftarrow 0$
}
  
  $\textit{Earlier} = \left\{ \hat{x} \in \calT \cup \calC \, \middle| \, \langle v, \hat{x} \rangle \in \calA \wedge  \hat{x}^{end} \leq x^{start} \right\}$ 

 \If{$\textit{Earlier} \neq \emptyset $}
 {
   $x_{prev} = \argmax_{\hat{x} \, \in \, \textit{Earlier}} \hat{x}^{end}$
   
    $m_{prev} \leftarrow T\left(x_{prev}^{destination}, x^{origin}\right)$
    
    $\textit{cost} \leftarrow \textit{cost}  +    E(v, m_{prev}) + \alpha \cdot \left( x^{start} - x^{end}_{prev} \right)$
 }
 
   $\textit{Later} = \left\{ \hat{x} \in \calT \cup \calC \, \middle| \, \langle v, \hat{x} \rangle \in \calA \wedge  x^{end} \leq \hat{x}^{start} \right\}$  
   
 \If{$\textit{Later} \neq \emptyset $}{
 $x_{next} = \argmin_{\hat{x} \, \in \, \textit{Earlier}} \hat{x}^{start}$
 
    $m_{next} \leftarrow T\left(x^{destination}, x_{next}^{origin}\right)$
    
    $\textit{cost} \leftarrow \textit{cost} +   E(v,m_{next}) + \alpha \cdot \left( x^{end} - x^{start}_{next} \right)$
 }
\KwResult{$\textit{cost}$}
\end{algorithm}

\paragraph{Biased Energy Cost}

Our greedy approach uses \cref{algo:energy_cost} to compute a \emph{biased energy cost} of assigning a bus~$v$ to a transit trip or charging slot $x$. If $x$ is a transit trip (i.e., $x \in \calT$), then the base cost of the assignment is $E(v, x)$. If $x$ is a charging slot, then the base cost is zero.
To compute the actual cost, the algorithm checks if bus $v$ is already assigned to any earlier (or later) transit trips or charging slots.
If it is, then it factors in the cost of the moving trip $m_{prev}$ (and $m_{next}$) from the preceding (and to the following) assignment $x_{prev}$ (and $x_{next}$).
Finally, the algorithm adds a bias to the actual cost based on the waiting time between $x_{prev}^{end}$ and $x^{start}$ (if $x_{prev}$ exists) and between $x^{end}$ and $x_{next}^{end}$ (if $x_{next}$ exists).
By adding these waiting times to the cost with an appropriate factor $\alpha > 0$, we nudge the greedy selection towards increasing bus utilization and minimizing layovers.
%The time complexity of this algorithm is $\mathcal{O}\left(|\mathcal{T} \cup \mathcal{C}|\right)$.

\begin{algorithm}[t]
 \caption{$\textbf{Greedy}(\mathcal{V}, \mathcal{T}, \mathcal{C}, \alpha)$}
 \label{algo:greedy_approach}
 
 $\mathcal{A} \leftarrow \emptyset$

$\calE \leftarrow \left\{ \langle v, t \rangle \mapsto \textbf{BiasedCost}(\calA, v, t, \alpha) ~ | ~ v \in \calV, t \in \calT \right\} $
 
 \While {$|\mathcal{T}| > 0$ \textnormal{\textbf{and}} $\min [ \calE ] \neq \infty$}
 {  
           $\textit{MinimumCostAssignments} \leftarrow \text{argmin}(\calE)$
          
         $\langle v, t \rangle \leftarrow \textbf{first}(\textit{MinimumCostAssignments})$
                 
        $\mathcal{A} \leftarrow \mathcal{A} \cup \{ \langle v,t \rangle \}$ 
        
        $\mathcal{T} \leftarrow \mathcal{T} \setminus \{t\}$
        
        $\calE, \calA \leftarrow \textbf{Update}(\calA, \mathcal{T}, \mathcal{C}, \calE, v, t, \alpha)$

        \Comment{update cost values $\calE$ and and add charging slots to the assignments as necessary}
}
 \KwResult{$\mathcal{A}$}
\end{algorithm}

\cref{algo:greedy_approach} shows our iterative greedy approach for assigning transit trips and charging slots to buses. The algorithm begins by computing the biased assignment cost for each pair of a bus $v$ and transit trip $t$ using $\textbf{BiasedCost}(\mathcal{A}, v, t, \alpha)$.
Starting with an empty set $\calA = \emptyset$,
the algorithm then iteratively adds assignments $(v, t) \in \calV \times \calT$ to the set, always choosing an assignment with the lowest biased cost $\textnormal{\textbf{BiasedCost}}(\calA, v, t, \alpha)$ (breaking ties arbitrarily).
After each iteration, the biased costs $\calE$ for the chosen vehicle $v$ are updated by $\textbf{Update}$, which also adds charging slot assignments as necessary (\sidenote{need to revise}see \cite{sivagnanam2021minimizing}). 
The algorithm terminates once all trips are assigned (or if it fails to find a solution).
The time complexity of \textbf{Update} is   $\mathcal{O}\left(|\calT|\cdot|\calV| + |\calT|\cdot|\calC|\cdot|\calX|\ln|\calX| \right)$,  where $\calX = \calT \cup \calC$. Since typically $|\calT| \gg |\calV|$, the complexity can be simplify into $\mathcal{O}\left(|\calT|\cdot|\calC|\cdot|\calX|\ln|\calX| \right)$.
Accordingly, the time complexity of the greedy algorithm is $\mathcal{O}\left(|\calT|^2\cdot|\calC|\cdot|\calX|\ln|\calX| \right)$. 
\end{copypasted}

Providing affordable public transit services is crucial for communities to ensure residents carry out daily activities such as employment, education, and other services without interruption. Unfortunately, transit agencies are facing inefficiency in utilizing transit resources, which results in high operating costs and environmental impact. Further, replacing traditional internal-combustion engine vehicles (ICEVs) with electric vehicles (EVs) can reduce energy costs and environmental impact. But due to the high upfront costs of EVs, transit agencies tend to use mixed fleets of vehicles (i.e. EVs and ICEVs).  To efficiently utilize the mixed fleets of vehicles, transit agencies must optimize how they assign each vehicle to serve a transit trip and ensure EVs are charged in a timely manner for uninterrupted services, which is a challenging problem for transit agencies with large transit networks~\cite{sivagnanam2021minimizing}. 
Because the advantage of using EVs inplace of ICEVs differs based on the route and time of the day. For example, EVs can be good at more traffic with frequent stops compared to be used on highways. Further, transit agencies have limited charging capabilities (i.e. number of charging poles, maximum power to consume at charging station to avoid high peak loads on the electric grid) which makes the problem more challenging.

\subsubsection{Problem Formulation}
We begin the mathematical formulation by defining a transit network, which includes a set of locations $\mathcal{L}$ such as bus stops, garages, and charging stations. Transit agency operates a set of \emph{buses}~$\mathcal{V}$, with the assistance of the transit network to serve the day-to-day transit services. Each bus $v \in \mathcal{V}$ belongs to a \emph{vehicle model}~$M_v \in \mathcal{M}$, and $\mathcal{M}$ is the set of all vehicle models in operation. Further, we divide vehicle models $\mathcal{M}$ into two disjoint subsets based on the energy source as liquid-fuel models $\mathcal{M}^\text{gas}$ (e.g. diesel, hybrid) and electric models $\mathcal{M}^\text{elec}$. Electric model ($m \in \mathcal{M}^\text{elec}$) has limited battery capacity ($C_m$) and may require charging during operational hours. On the other hand, liquid-fuel models can operate without refueling throughout the day. Every day the transit agencies must serve a set of \emph{transit trips}~$\mathcal{T}$ using its buses. Each trip $t \in \mathcal{T}$ starts from the origin $t^\text{origin} \in \mathcal{L}$ at $t^{\text{start}}$, then passes through a set of stops at fixed arrival and departure times, and finally reaches the destination $t^\text{destination} \in \mathcal{L}$ at $t^{\text{end}}$. 

The transit agency installs a set of \emph{charging poles} $\mathcal{CP}$ (generally at the depot) to charge electric buses. To better facilitate the operation of each charging pole $cp \in \mathcal{CP}$, the day can be divided into uniform-length disjoint \emph{time slots} $\mathcal{S}$, where each slot $s \in \mathcal{S}$ begins at~$s^{\text{start}}$ and ends at~$s^{\text{end}}$. We combine the set of charging poles and slots and obtain the charging slots $\mathcal{CS} = \mathcal{CP} \times \mathcal{S}$. Each electric bus ($v \in \mathcal{V} \land M_v \in \mathcal{M}^\text{elec}$) can be charged at any unassigned charging slot ($cs \in \mathcal{CS}$) at a time, as long the electric bus $v$ can reach the charging pole without running out of charge in the battery.

In between two transit trips $x_1, x_2 \in \mathcal{T} \bigcup \mathcal{CS}$ (or between a transit trip and a charging slot, or between two charging slots for electric buses), the bus may be required to move from destination for the previous trip $x_1$ to the origin of the following trip $x_2$. Such trips can be defined as \emph{non-service trips} since they are not directly contributing to serving passengers but are required for the operation of the buses throughout the day. Each on-service trip can be defined by $T(l_1, l_2)$ where the trips start at the location $l_1 \in \mathcal{L}$ and end at the location $l_2 \in \mathcal{L}$.

% need to summarize more
\paragraph{Solution} We define the solution as a set of trip assignments where each transit trip is assigned to exactly one bus $v \in \mathcal{V}$ and a set of charging assignments of the electric bus to the combination of a charging pole $cp \in \mathcal{CP}$ and specific slot $s \in \mathcal{S}$ on the charging pole.

\paragraph{Constraints} The solution must respect the following real-world constraints: Every bus can serve only one trip at a time; A bus can serve two trips consecutively (or an electric bus can serve a trip and assign to a charging), if the bus has enough time to move from the destination of the first trip to origin of the second trip before the start time of the second trip;  Every electric bus needs to have enough energy before serving every assigned trip; Only one electric bus can be charged in particular charging pole at a particular slot; the energy charged by electric bus is capped at the maximum capacity of the battery.

\paragraph{Objective} In the energy-efficient vehicle assignment, we aim to reduce the energy use of transit vehicles. We express our objective as 
\begin{equation*}
\min_{\calA} ~~\quad \sum_{\mathclap{v \in \calV: \, M_v \in \calM^\text{gas}}} ~ K^\text{gas} \! \cdot e(\calA, v) + ~~\quad \sum_{\mathclap{v \in \calV: \, M_v \in \calM^\text{elec}}} ~ K^\text{elec} \! \cdot e(\calA, v)
% \label{equ:obj_main}
\end{equation*}
where $e(\calA, v)$ be the amount of energy used by bus~$v$ for all trips completed at the end of the day and $K^\text{gas}$ and $K^\text{elec}$ denote the unit costs of energy used for liquid-fuel and electric vehicles

\subsubsection{Computational Approaches}

One way to solve the above problem is using an integer program (IP). Although the IP provides the optimal solution for smaller problem instances, the problem becomes computationally intractable for larger problem instances. In this sub-section, we introduce a greedy (to provide an initial solution) and simulated annealing-based metaheuristics approach that solves the larger problem instances in polynomial time.

\paragraph{Greedy Algorithm} In this algorithm, we follow an iterative process where we greedily choose an assignment between all possible assignments based on assigning a bus~$v$ to a transit trip or charging slot $x$ instead of the base energy cost to serve the request. Biased cost comprises three main costs (i) base energy cost (base energy cost for charging slot is 0),  (ii) the cost associated with moving the bus from the destination of the previous trip to the origin of the current trip, or vice versa, (iii) the wait time between the end of the previous trip to the beginning of the current trip or vice versa. The algorithm at the beginning computes the biased cost for the pairs of vehicles and trips. Then follows the iterative process, choosing the assignment with the lowest biased cost in each iteration. Then the algorithm updates the costs of remaining unassigned trips with respect to the vehicle used in the previous assignment. The algorithm follows the iterative process until all the requests are assigned or fail to find a solution.

\paragraph{Simulated Annealing Algorithm} The algorithm starts with the solution from the greedy algorithm and performs an iterative random search where the algorithm generates a random neighbor to the current solution in each iteration. Generating the random neighbor follows an iterative process where the algorithm randomly chooses the two vehicles in each iteration, splits their operational hours into two halves, and tries to swap the assignment between two vehicles. The algorithm to generate a random neighbor terminates after enough swapping iterations are performed. Since simulated annealing is an anytime algorithm, it can be configured to run until a fixed time.

\subsubsection{Numerical Results}

\colorlet{ColorSimAnn}{magenta}
\colorlet{ColorLegendSimAnn}{ColorSimAnn!50}
\colorlet{ColorGreedy}{blue}
\colorlet{ColorLegendGreedy}{ColorGreedy!50}
\colorlet{ColorReal}{black}
\colorlet{ColorLegendReal}{ColorReal!50}

This section discusses the numerical results using the computational approach presented in the earlier sections. First, we obtain the GTFS, vehicles, and charging pole data from a real-transit agency (i.e. Chattanooga Area Regional Transportation Authority (CARTA)). To evaluate the algorithm's effectiveness presented earlier, we obtain real-world GTFS data from CARTA. Since routes for the non-service trips are not specified as part of GTFS, we obtain the routes using Google Directions API. Based on the GTFS data and routes obtained from Google Directions API, we compute the energy consumption for serving the transit and non-service trips using EVs and ICEVs based on the energy estimation described in the earlier \cref{subsec:energy_prediction}. After that, we feed the energy optimization into our computational approaches and run experiments. 

\pgfplotstableread[col sep=comma,]{energy_assignment/data/greedy_diff_costs.csv}\greedycost
\pgfplotstableread[col sep=comma,]{energy_assignment/data/sim_anneal_diff_costs.csv}\simannealcost

\begin{figure}
\begin{tikzpicture}
\begin{axis}[
      boxplot/draw direction=y,
      xtick={1,2,3,4,5,6,7,8},
      width=0.75*\columnwidth,
      height = 4.5cm,
      font=\small,
      bugsResolvedStyle/.style={},
      ylabel={Energy Cost},
      xlabel={Number of Bus Lines},
      ymin=95,
      ymax=225,
      yticklabel=\pgfmathprintnumber{\tick}\,$\%$,
      legend pos=north west,
      legend columns=2,
      legend cell align={left},
    ]
    
    \addplot+[boxplot={box extend=0.25, draw position=1},ColorGreedy, rshift, solid, fill=ColorGreedy!20, mark=+] table [col sep=comma, y=line_1] {\greedycost};
\addplot+[boxplot={box extend=0.25, draw position=2},ColorGreedy, rshift, solid, fill=ColorGreedy!20, mark=+] table [col sep=comma, y=line_2] {\greedycost};
\addplot+[boxplot={box extend=0.25, draw position=3},ColorGreedy, rshift, solid, fill=ColorGreedy!20, mark=+] table [col sep=comma, y=line_3] {\greedycost};
\addplot+[boxplot={box extend=0.25, draw position=4},ColorGreedy, rshift, solid, fill=ColorGreedy!20, mark=+] table [col sep=comma, y=line_4] {\greedycost};
\addplot+[boxplot={box extend=0.25, draw position=5},ColorGreedy, rshift, solid, fill=ColorGreedy!20, mark=+] table [col sep=comma, y=line_5] {\greedycost};
\addplot+[boxplot={box extend=0.25, draw position=6},ColorGreedy, rshift, solid, fill=ColorGreedy!20, mark=+] table [col sep=comma, y=line_6] {\greedycost};
\addplot+[boxplot={box extend=0.25, draw position=7},ColorGreedy, rshift, solid, fill=ColorGreedy!20, mark=+] table [col sep=comma, y=line_7] {\greedycost};
\addplot+[boxplot={box extend=0.25, draw position=8},ColorGreedy, rshift, solid, fill=ColorGreedy!20, mark=+] table [col sep=comma, y=line_8] {\greedycost};
\addplot+[lshift, boxplot={box extend=0.25, draw position=1}, ColorSimAnn, solid, fill=ColorSimAnn!20, mark=x] table [col sep=comma, y=line_1] {\simannealcost};
\addplot+[lshift, boxplot={box extend=0.25, draw position=2}, ColorSimAnn, solid, fill=ColorSimAnn!20, mark=x] table [col sep=comma, y=line_2] {\simannealcost};
\addplot+[lshift, boxplot={box extend=0.25, draw position=3}, ColorSimAnn, solid, fill=ColorSimAnn!20, mark=x] table [col sep=comma, y=line_3] {\simannealcost};
\addplot+[lshift, boxplot={box extend=0.25, draw position=4}, ColorSimAnn, solid, fill=ColorSimAnn!20, mark=x] table [col sep=comma, y=line_4] {\simannealcost};
\addplot+[lshift, boxplot={box extend=0.25, draw position=5}, ColorSimAnn, solid, fill=ColorSimAnn!20, mark=x] table [col sep=comma, y=line_5] {\simannealcost};
\addplot+[lshift, boxplot={box extend=0.25, draw position=6}, ColorSimAnn, solid, fill=ColorSimAnn!20, mark=x] table [col sep=comma, y=line_6] {\simannealcost};
\addplot+[lshift, boxplot={box extend=0.25, draw position=7}, ColorSimAnn, solid, fill=ColorSimAnn!20, mark=x] table [col sep=comma, y=line_7] {\simannealcost};
\addplot+[lshift, boxplot={box extend=0.25, draw position=8}, ColorSimAnn, solid, fill=ColorSimAnn!20, mark=x] table [col sep=comma, y=line_8] {\simannealcost};
\end{axis}
\end{tikzpicture}
\caption{Energy cost for operating transit services based on the assignment obtained from simulated annealing (\textcolor{ColorLegendSimAnn}{$\blacksquare$}) and the greedy algorithm~(\textcolor{ColorLegendGreedy}{$\blacksquare$}) compared to optimal assignments using IP.}
	\label{fig:cost_all}
\end{figure}
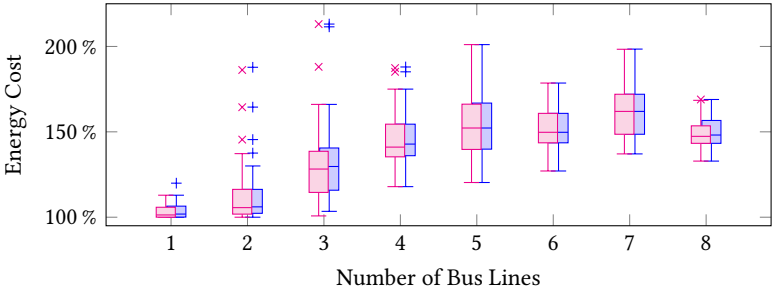

% not sure how much detail I need to provide numerical results

% need to revise
\paragraph{Solution quality} Based on the experiments on how well our heuristic and meta-heuristic minimize the energy cost compared to the Integer program, we observe that our solution approaches can restrict the solution quality to be within the range of 1.5 to 1.6, even for larger instances. For reference, you can look at the \cref{fig:cost_all}, which provides the energy cost of the solution obtained by our heuristic approaches compared to IP in percentages when serving 1-12 bus lines with exactly 10 trips using 3 EVs and ICEVs 5 times as the number of bus-lines capped at 50 ICEVs (i.e. for the scenario with 11 and 12 bus lines we only consider 50 ICEVs)

%\subsection{Practical Considerations}

%- data management

%- privacy

%- computation availability and management

%- operational limitation : failures, drivers, vehicle schedules

\section{Future Considerations}

Efficient transportation systems require making decisions in real-time in complex environments.
Historically, public transit systems rely on human intuition and analysis to design fixed schedules ahead of time.
New demand-responsive modalities such as ride-share, microtransit, and paratransit use control algorithms to match trip requests to vehicles in real-time to provide on-demand service.
Next generation systems increasingly aim to leverage AI to automate and increase efficiency across modalities using data.
In this chapter, we discussed methodologies for leveraging AI in transportation and provided some concrete examples of computational problems in this space.

A major consideration for transit agencies is to increase utilization of fixed-line public transit. 
Public transit is highly efficient in terms of energy and emissions per passenger particularly compared to private cars and ride-share.
Therefore, to combat climate change it is important to continue investigating ways to increase ridership on fixed-line services.
One important future consideration is the potential of multi-modal, hybrid transportation systems that combine on-demand systems with fixed-line service in a way that uses on-demand mobility to serve the first-mile-last-mile (FMLM) to and from fixed-line.
These systems have the potential to combine the best parts of demand-responsive and fixed-line services in a way that increases fixed-line utilization by expanding access to more residents.

Autonomy in general has great potential to improve mobility options in urban environments.
We already discussed autonomy at the system level, e.g. in the context of designing real-time ride-share and ride-pooling algorithms or optimizing trip-vehicle assignment for fixed-line.
Moving forward, there is great potential for improving control systems for autonomous vehicles.
At a system level, autonomous vehicles have the potential to reduce accidents as well as traffic, thereby increasing safety and efficiency~\cite{lichtle2022deploying, stern2018dissipation}.
As these systems become more capable it will be important for transit agencies to adapt to solve increasingly complicated control and optimization problems that arise from operating mixed fleets that include conventional as well electric and autonomous vehicles.

\section*{Acknowledgements}

The authors will like to acknowledge the collaborative efforts of Dr. Ayan Mukhopadhyay, Vanderbilt University and Mr. Philip Pugliese from Chattanooga  Area Regional Transportation Authority. The work presented here has been sponsored in part by the National Science Foundation through 1952011 and Department of Energy through awards DE-EE0008467 and DE-EE0009212. Any opinions, findings, and conclusions or recommendations expressed in this material are those of the author(s) and do not necessarily reflect the views of the National Science Foundation and Department of Energy.

\bibliographystyle{plain}
\bibliography{main.bib}

\begin{thebibliography}{10}

\bibitem{alonso2017demand}
Javier Alonso-Mora, Samitha Samaranayake, Alex Wallar, Emilio Frazzoli, and
  Daniela Rus.
\newblock On-demand high-capacity ride-sharing via dynamic trip-vehicle
  assignment.
\newblock {\em Proceedings of the National Academy of Sciences},
  114(3):462--467, 2017.

\bibitem{arulkumaran2017deep}
Kai Arulkumaran, Marc~Peter Deisenroth, Miles Brundage, and Anil~Anthony
  Bharath.
\newblock Deep reinforcement learning: A brief survey.
\newblock {\em IEEE Signal Processing Magazine}, 34(6):26--38, 2017.

\bibitem{aydin2016modeling}
Berkay Aydin, Vijay Akkineni, and Rafal~A Angryk.
\newblock Modeling and indexing spatiotemporal trajectory data in
  non-relational databases.
\newblock In {\em Managing Big Data in Cloud Computing Environments}, pages
  133--162. IGI Global, 2016.

\bibitem{ayman2022data}
Afiya Ayman, Amutheezan Sivagnanam, Michael Wilbur, Philip Pugliese, Abhishek
  Dubey, and Aron Laszka.
\newblock Data-driven prediction and optimization of energy use for transit
  fleets of electric and ice vehicles.
\newblock {\em ACM Transactions on Internet Technology}, 22(1):7:1--7:29,
  February 2022.

\bibitem{ayman2020data}
Afiya Ayman, Michael Wilbur, Amutheezan Sivagnanam, Philip Pugliese, Abhishek
  Dubey, and Aron Laszka.
\newblock Data-driven prediction of route-level energy use for mixed-vehicle
  transit fleets.
\newblock In {\em 6th IEEE International Conference on Smart Computing
  (SMARTCOMP)}, pages 41--48, September 2020.

\bibitem{beyazit2011evaluating}
Eda Beyazit.
\newblock Evaluating social justice in transport: lessons to be learned from
  the capability approach.
\newblock {\em Transport Reviews}, 31(1):117--134, 2011.

\bibitem{lodesdata}
The United States~Census Bureau.
\newblock Longitudinal employer-household dynamics, 2023.

\bibitem{chen2020review}
Yuche Chen, Guoyuan Wu, Ruixiao Sun, Abhishek Dubey, Aron Laszka, and Philip
  Pugliese.
\newblock A review and outlook on energy consumption estimation models for
  electric vehicles.
\newblock {\em SAE International Journal of Sustainable Transportation, Energy,
  Environment, \& Policy}, 2(1):79--96, March 2021.

\bibitem{chollet2015keras}
Fran\c{c}ois Chollet et~al.
\newblock Keras.
\newblock \url{https://keras.io}, 2015.

\bibitem{eia}
{EIA}.
\newblock {U.S.} {Energy} {Information} {Administration}: Use of energy
  explained -- energy use for transportation (2018).
\newblock
  \url{https://www.eia.gov/energyexplained/use-of-energy/transportation.php},
  Accessed: May 31st, 2020, 2018.

\bibitem{federal2003status}
{Federal Highway Administration}.
\newblock Status of the nation’s highways, bridges, and transit: 2002
  conditions and performance report, 2003.

\bibitem{realtimegtfs}
Google.
\newblock General transit feed specification ({GTFS}) real-time overview, 2022.

\bibitem{staticgtfs}
Google.
\newblock General transit feed specification ({GTFS}) static overview, 2022.

\bibitem{haklay2008openstreetmap}
Mordechai Haklay and Patrick Weber.
\newblock {OpenStreetMap}: User-generated street maps.
\newblock {\em IEEE Pervasive Computing}, 7(4):12--18, 2008.

\bibitem{harvey2010social}
David Harvey.
\newblock {\em Social justice and the city}, volume~1.
\newblock University of Georgia press, 2010.

\bibitem{inrixdata}
INRIX.
\newblock Inrix traffic, 2022.

\bibitem{joe2020deep}
Waldy Joe and Hoong~Chuin Lau.
\newblock Deep reinforcement learning approach to solve dynamic vehicle routing
  problem with stochastic customers.
\newblock In {\em Conference on Automated Planning and Scheduling (ICAPS)},
  volume~30, pages 394--402, 2020.

\bibitem{kingma2014adam}
Diederik~P Kingma and Jimmy Ba.
\newblock Adam: A method for stochastic optimization.
\newblock {\em arXiv preprint arXiv:1412.6980}, 2014.

\bibitem{lichtle2022deploying}
Nathan Lichtl{\'e}, Eugene Vinitsky, Matthew Nice, Benjamin Seibold, Dan Work,
  and Alexandre~M Bayen.
\newblock Deploying traffic smoothing cruise controllers learned from
  trajectory data.
\newblock In {\em 2022 International Conference on Robotics and Automation
  (ICRA)}, pages 2884--2890. IEEE, 2022.

\bibitem{valhalla}
MapZen.
\newblock Valhalla, 2023.

\bibitem{meteostat}
Meteostat.
\newblock Historical weather and climate data.
\newblock \url{https://meteostat.net/en} Raw data: NOAA, Deutscher
  Wetterdienst, 2020.

\bibitem{moerland2023model}
Thomas~M Moerland, Joost Broekens, Aske Plaat, Catholijn~M Jonker, et~al.
\newblock Model-based reinforcement learning: A survey.
\newblock {\em Foundations and Trends in Machine Learning}, 16(1):1--118, 2023.

\bibitem{mohammadi2018deep}
Mehdi Mohammadi, Ala Al-Fuqaha, Sameh Sorour, and Mohsen Guizani.
\newblock Deep learning for iot big data and streaming analytics: A survey.
\newblock {\em IEEE Communications Surveys \& Tutorials}, 20(4):2923--2960,
  2018.

\bibitem{EPA-420-F-19-047}
{Office of Transportation and Air Quality}.
\newblock Fast facts: {U.S.} transportation sector greenhouse gas emissions
  1990--2017.
\newblock Technical Report EPA-420-F-19-047, U.S. Environmental Protection
  Agency, June 2019.

\bibitem{osrm}
Project-OSRM.
\newblock Open source routing machine, 2022.

\bibitem{puterman2014markov}
Martin~L Puterman.
\newblock {\em Markov decision processes: discrete stochastic dynamic
  programming}.
\newblock John Wiley \& Sons, 2014.

\bibitem{safegraph}
SafeGraph.
\newblock Safegraph data documentation, 2023.

\bibitem{salazar2018interaction}
Mauro Salazar, Federico Rossi, Maximilian Schiffer, Christopher~H Onder, and
  Marco Pavone.
\newblock On the interaction between autonomous mobility-on-demand and public
  transportation systems.
\newblock In {\em 2018 21st International Conference on Intelligent
  Transportation Systems (ITSC)}, pages 2262--2269. IEEE, 2018.

\bibitem{dtr}
{Scikit-learn Developers}.
\newblock Sklearn.tree.decisiontreeregressor.
\newblock
  \url{https://scikit-learn.org/stable/modules/generated/sklearn.tree.DecisionTreeRegressor.html},
  Accessed: May 31st, 2020.

\bibitem{sejourne2018price}
Thibault S{\'e}journ{\'e}, Samitha Samaranayake, and Siddhartha Banerjee.
\newblock The price of fragmentation in mobility-on-demand services.
\newblock {\em Proceedings of the ACM on Measurement and Analysis of Computing
  Systems}, 2(2):1--26, 2018.

\bibitem{sivagnanam2021minimizing}
Amutheezan Sivagnanam, Afiya Ayman, Michael Wilbur, Philip Pugliese, Abhishek
  Dubey, and Aron Laszka.
\newblock Minimizing energy use of mixed-fleet public transit for fixed-route
  service.
\newblock In {\em 35th AAAI Conference on Artificial Intelligence (AAAI)},
  pages 14930--14938, February 2021.

\bibitem{sivagnanam2022offline}
Amutheezan Sivagnanam, Salah~Uddin Kadir, Ayan Mukhopadhyay, Philip Pugliese,
  Abhishek Dubey, Samitha Samaranayake, and Aron Laszka.
\newblock Offline vehicle routing problem with online bookings: A novel problem
  formulation with applications to paratransit.
\newblock In {\em 31st International Joint Conference on Artificial
  Intelligence (IJCAI)}, pages 3933--3939, July 2022.

\bibitem{stern2018dissipation}
Raphael~E Stern, Shumo Cui, Maria~Laura Delle~Monache, Rahul Bhadani, Matt
  Bunting, Miles Churchill, Nathaniel Hamilton, Hannah Pohlmann, Fangyu Wu,
  Benedetto Piccoli, et~al.
\newblock Dissipation of stop-and-go waves via control of autonomous vehicles:
  Field experiments.
\newblock {\em Transportation Research Part C: Emerging Technologies},
  89:205--221, 2018.

\bibitem{heredata}
HERE Technologies.
\newblock Here traffic, 2022.

\bibitem{geofabrik}
Jochen Topf and Frederik Ramm.
\newblock Geofabrik, 2020.

\bibitem{wang2019integrated}
Shaohua Wang, Yang Zhong, and Erqi Wang.
\newblock An integrated gis platform architecture for spatiotemporal big data.
\newblock {\em Future Generation Computer Systems}, 94:160--172, 2019.

\bibitem{wilbur2020impact}
Michael Wilbur, Afiya Ayman, Amutheezan Sivagnanam, Anna Ouyang, Vincent Poon,
  Riyan Kabir, Abhiram Vadali, Philip Pugliese, Daniel Freudberg, Aron Laszka,
  et~al.
\newblock Impact of covid-19 on public transit accessibility and ridership.
\newblock {\em Transportation Research Record}, page 03611981231160531, 2020.

\bibitem{wilbur2022iccps}
Michael Wilbur, Salah~Uddin Kadir, Youngseo Kim, Geoffrey Pettet, Ayan
  Mukhopadhyay, Philip Pugliese, Samitha Samaranayake, Aron Laszka, and
  Abhishek Dubey.
\newblock An online approach to solve the dynamic vehicle routing problem with
  stochastic trip requests for paratransit services.
\newblock In {\em 13th ACM/IEEE International Conference on Cyber-Physical
  Systems (ICCPS)}, pages 147--158, May 2022.

\bibitem{ecml2021}
Michael Wilbur, Ayan Mukhopadhyay, Sayyed Vazirizade, Philip Pugliese, Aron
  Laszka, and Abhishek Dubey.
\newblock Energy and emission prediction for mixed-vehicle transit fleets using
  multi-task and inductive transfer learning.
\newblock In {\em 2021 European Conference on Machine Learning and Principles
  and Practice of Knowledge Discovery in Databases (ECML PKDD)}, pages
  502--517, September 2021.

\bibitem{wilbur2021efficient}
Michael Wilbur, Philip Pugliese, Aron Laszka, and Abhishek Dubey.
\newblock Efficient data management for intelligent urban mobility systems.
\newblock In {\em Proceedings of the Workshop on Data-Driven and Intelligent
  Cyber-Physical Systems (DI-CPS'21)}, pages 22--26, May 2021.

\end{thebibliography}

\end{document}